\documentclass{article}

\usepackage{microtype}
\usepackage{graphicx}
\usepackage{booktabs}
\usepackage{multirow}
\usepackage{amsmath}
\usepackage{amssymb}
\usepackage{amsthm}
\usepackage{xcolor}
\usepackage{pifont}
\usepackage{tikz}
\usetikzlibrary{arrows.meta,positioning,fit,calc}
\usepackage{hyperref}

\usepackage[accepted]{mlsys2025}
\makeatletter\renewcommand{\mlsys@appearing}{Preprint.}\makeatother
\hypersetup{pdftitle={Scute: Exact Batched Serving for PEFT Adapters Beyond LoRA}, pdfauthor={Thomas Walker, Richard Baraniuk}}

\newtheorem{proposition}{Proposition}
\newtheorem{corollary}{Corollary}

\newcommand{\dout}{d_{\mathrm{out}}}
\newcommand{\dinn}{d_{\mathrm{in}}}

\usepackage{bm}
\usepackage{pgffor}
\makeatletter
\foreach \x in {a,b,c,d,e,g,h,i,j,k,l,n,o,p,q,r,s,t,u,v,w,x,y,z} {
    \expandafter\xdef\csname b\x\endcsname{\noexpand\ensuremath{\noexpand\boldsymbol{\x}}}
}
\foreach \x in {A,...,Z} {
    \expandafter\xdef\csname b\x\endcsname{\noexpand\ensuremath{\noexpand\boldsymbol{\x}}}
}
\makeatother
\newcommand{\bmm}{\ensuremath{\boldsymbol{m}}}

\newcommand{\nBeta}{680}
\newcommand{\nBetaLtwo}{3205}
\newcommand{\nBmCellsGather}{27}
\newcommand{\nBmCellsIndex}{27}
\newcommand{\nBmCellsOutside}{28}
\newcommand{\nBmCellsSide}{27}
\newcommand{\nBmErrGather}{3.4\%}
\newcommand{\nBmErrIndex}{14.0\%}
\newcommand{\nBmErrLaunchGather}{23\%}
\newcommand{\nBmErrLaunchIndex}{24\%}
\newcommand{\nBmErrLaunchOutside}{21\%}
\newcommand{\nBmErrLaunchSide}{2\%}
\newcommand{\nBmErrOutside}{0.7\%}
\newcommand{\nBmErrSide}{1.4\%}
\newcommand{\nBmIncGather}{7.4\%}

\newcommand{\nBmIncIndex}{20.0\%}

\newcommand{\nBmIncOutside}{1.2\%}
\newcommand{\nBmIncSide}{3.2\%}

\newcommand{\nDefaultQkvRoofHi}{60\%}
\newcommand{\nDefaultQkvRoofLo}{58\%}

\newcommand{\nEngOutHi}{2.95$\times$}
\newcommand{\nEngOutLo}{2.38$\times$}

\newcommand{\nEngSlotHi}{1.022$\times$}
\newcommand{\nEngSlotLo}{0.985$\times$}
\newcommand{\nFoldFullRank}{60 of 60}
\newcommand{\nFoldGdev}{0.2\%}
\newcommand{\nFoldLeftSixteen}{59.7\%}

\newcommand{\nLambda}{0.96}

\newcommand{\nMeasSlotDecRatio}{1.002$\times$}

\newcommand{\nOftCells}{120}

\newcommand{\nOftIncErr}{12.3\%}

\newcommand{\nOftPredErr}{1.2\%}

\newcommand{\nTpFoldComparisons}{3328}

\newcommand{\hAfISixteen}{1.34$\times$}

\newcommand{\hAfMixDecHi}{1.69$\times$}
\newcommand{\hAfMixDecLo}{1.12$\times$}

\newcommand{\hAfMixPreLo}{0.76$\times$}

\newcommand{\hAfOSixteen}{1.31$\times$}

\newcommand{\hAfRefHi}{1.89$\times$}
\newcommand{\hAfRefLo}{1.07$\times$}
\newcommand{\hAfVSixteen}{1.83$\times$}
\newcommand{\hAgreeAdapters}{9}
\newcommand{\hAgreeBest}{0.3\%}
\newcommand{\hAgreeWorst}{1.5\%}
\newcommand{\hAhAgreeHi}{1.6\%}
\newcommand{\hAhAgreeLo}{0.3\%}
\newcommand{\hBytesSixtyFour}{0.34}

\newcommand{\hCapASixtyFourSustainOurs}{20}
\newcommand{\hCapASixtyFourSustainStock}{8}

\newcommand{\hCapAThousandSloLowStock}{32\%}
\newcommand{\hCapAThousandSustainOurs}{8}

\newcommand{\hCapAThrHi}{1.8$\times$}
\newcommand{\hCapAThrLo}{1.2$\times$}
\newcommand{\hCapATopRate}{24}

\newcommand{\hCapATwoFiftySixSustainOurs}{8}

\newcommand{\hCapOursResident}{1024}
\newcommand{\hCapOursResidentKv}{98k}

\newcommand{\hCapSixtyFourSustainOurs}{10}

\newcommand{\hCapStockFortyEightKv}{75k}
\newcommand{\hCapStockResidentKv}{28k}
\newcommand{\hCapStockResidentMax}{64}
\newcommand{\hCapSustainHi}{2.5$\times$}
\newcommand{\hCapSustainLo}{1.5$\times$}
\newcommand{\hCapThousandSloOurs}{82\%}
\newcommand{\hCapThousandSloStock}{9\%}
\newcommand{\hCapThousandSustainOurs}{6}
\newcommand{\hCapThousandSustainRatio}{1.5$\times$}
\newcommand{\hCapThousandSustainStock}{4}
\newcommand{\hCapThousandThr}{2.01$\times$}
\newcommand{\hCapThousandThrTen}{2.03$\times$}
\newcommand{\hCapThousandTokOurs}{932}
\newcommand{\hCapThousandTokStock}{464}
\newcommand{\hCapThousandTtft}{65$\times$}

\newcommand{\hCapTwoFiftySixSustainOurs}{8}

\newcommand{\hConvAsvdGenBelow}{8}
\newcommand{\hConvAsvdGenWorstSixtyFour}{19.6}

\newcommand{\hConvAsvdOwnWorstSixtyFour}{1.2}

\newcommand{\hConvDistillWorstSixtyFour}{2.0}

\newcommand{\hConvSvdBelow}{9}
\newcommand{\hConvSvdWorstSixtyFour}{23.2}
\newcommand{\hConvVeraSvdWorstSixteen}{18.5}
\newcommand{\hConvVeraSvdWorstSixtyFour}{0.6}

\newcommand{\hFAddMStock}{34.7}

\newcommand{\hFDecMOwn}{13.0}
\newcommand{\hFDecMPad}{31.6}

\newcommand{\hFISixteen}{1.23$\times$}

\newcommand{\hFMMemHi}{44.9}
\newcommand{\hFMMemLo}{44.8}

\newcommand{\hFMPreHi}{5.96$\times$}

\newcommand{\hFOOursSlope}{0.16}

\newcommand{\hFOrderCells}{102}
\newcommand{\hFOrderGap}{0.4\%}

\newcommand{\hFPreOneHi}{1.04$\times$}
\newcommand{\hFPreOneLo}{1.02$\times$}
\newcommand{\hFPreSixteenHi}{1.09$\times$}
\newcommand{\hFPreSixteenLo}{1.05$\times$}

\newcommand{\hFVOursOne}{35.4}
\newcommand{\hFVOursSixteen}{36.1}
\newcommand{\hFVOursSlope}{0.05}
\newcommand{\hFVSixteen}{1.80$\times$}
\newcommand{\hFVStockOne}{36.7}
\newcommand{\hFVStockSixteen}{65.1}

\newcommand{\hQBaseBoolq}{74.0\%}
\newcommand{\hQBaseGsm}{26.2\%}
\newcommand{\hQBaseMnli}{36.8\%}

\newcommand{\hQualOursMaxDiff}{0.4 points}
\newcommand{\hSBaseFused}{30.6}

\newcommand{\hSBaseSplit}{26.8}
\newcommand{\hSDecRefHi}{17\%}
\newcommand{\hSDecRefLo}{5}
\newcommand{\hSMDec}{1.65$\times$}
\newcommand{\hSMDecRef}{1.17$\times$}

\newcommand{\hSMPreRef}{1.18$\times$}

\newcommand{\hSMnvDecRef}{1.05$\times$}

\newcommand{\hSMnvPreRef}{1.04$\times$}

\newcommand{\hSOSixteen}{1.23$\times$}
\newcommand{\hSOSixteenShipped}{1.34$\times$}
\newcommand{\hSPreRefHi}{18\%}
\newcommand{\hSPreRefLo}{4}

\newcommand{\hSVDec}{1.12$\times$}
\newcommand{\hSVDecPct}{12\%}

\newcommand{\hSVDecSixteen}{1.13$\times$}

\newcommand{\hShippedDecHi}{1.98$\times$}
\newcommand{\hShippedDecLo}{1.21$\times$}
\newcommand{\hSurveyAuthors}{10}
\newcommand{\hSurveyBelow}{144}
\newcommand{\hSurveyMax}{10.3\%}
\newcommand{\hSurveyMedian}{0.16\%}
\newcommand{\hSurveyN}{149}
\newcommand{\hTrainedDoraMedian}{0.22\%}

\mlsystitlerunning{Scute: Exact Batched Serving for PEFT Adapters Beyond LoRA}

\begin{document}

\twocolumn[
\mlsystitle{Scute: Exact Batched Serving for PEFT Adapters Beyond LoRA}

\begin{mlsysauthorlist}
\mlsysauthor{Thomas Walker}{rice}
\mlsysauthor{Richard Baraniuk}{rice}
\end{mlsysauthorlist}
\mlsysaffiliation{rice}{Rice University, Houston, TX, USA}
\mlsyscorrespondingauthor{Thomas Walker}{tw78@rice.edu}

\mlsyskeywords{LLM serving, parameter-efficient fine-tuning, LoRA, DoRA, VeRA, OFT, (IA)$^3$, GPU kernels}

\vskip 0.3in

\begin{abstract}
    Existing serving systems for large language models batch requests for many different LoRA adapters on one GPU.
    However, these systems reject adapters from other parameter-efficient fine-tuning (PEFT) methods such as DoRA, (IA)$^3$, VeRA and OFT.
    One solution to this is to convert these adapters to LoRA, but this induces a significant memory and accuracy cost.
    Instead, we show that, under some modifications, a batched LoRA kernel can serve all of these methods exactly and efficiently.
    We achieve this by writing each method in one normal form, a LoRA computation with six \emph{slots} for an adapter's own vectors and matrices.
    Some slots can merge into an adapter's weights at load time; the others remain live and induce a cost.
    We give ways to efficiently pay this cost and predict it from a zero-parameter byte model.
    We package this system, Scute, and build it on top of vLLM.
    Scute serves all five methods, coming as checkpoints from the Hugging Face PEFT library, in one batch, and it matches that library with a mean error of at most \hAgreeWorst{} of each adapter's effect.
    Under a burst of requests on an RTX A6000, a mix of these methods takes \hSDecRefLo--\hSDecRefHi{} more decode time than the same number of LoRA adapters.
    Furthermore, using Scute, the same GPU can serve 64 to 1024 (IA)$^3$ and VeRA adapters at \hCapSustainLo{} to at least \hCapSustainHi{} the request rate of their LoRA conversions.
\end{abstract}
]

\printAffiliationsAndNotice{}

\section{Introduction}
\label{sec:intro}

Parameter-efficient fine-tuning (PEFT) adapts a pretrained model to a task by training a small \emph{adapter} while the model's own weights stay frozen.
A platform that serves many customers, or \emph{tenants}, can then keep one copy of the base model on a GPU and apply each tenant's adapter on top of it.
LoRA~\citep{lora}, the most common method, adds a low-rank update to each adapted weight matrix, but other methods are also in use.
Most of the methods below are implemented in the Hugging Face PEFT library~\citep{peft}, which we call the \emph{PEFT library} and use as the reference implementation.
DoRA~\citep{dora} splits the update into a magnitude and a direction, (IA)$^3$~\citep{ia3} rescales activations, SSF~\citep{ssf} rescales and shifts them, VeRA~\citep{vera} shares two frozen random matrices across all adapters and trains only two small vectors per layer, and OFT~\citep{oft,boft} rotates the input of each layer.
Which method works best can depend on the setting, including for adapters that a hypernetwork generates at inference time~\citep{walker2026adapterbench}, so a platform may need to serve several methods side by side.

Most serving systems support only LoRA.
Punica~\citep{punica} batches requests for many LoRA adapters into one kernel launch, and S-LoRA~\citep{slora} pages thousands of LoRA adapters between host and GPU memory.
The production engines vLLM~\citep{pagedattention} and SGLang~\citep{sglang} use this design, and they reject adapters of other methods at load.\footnote{vLLM 0.32.0 and SGLang 0.5.22, release and \texttt{main}.
An open vLLM pull request adds DoRA (\S\ref{sec:related}).
TensorRT-LLM~\citep{trtllm} served DoRA in its TensorRT backend up to release 1.2, and the 1.3 and 1.4 release candidates do not have that backend.
We build on vLLM 0.19.1, the newest release that runs on the CUDA~12.2 driver of our machines.}

A platform can convert other adapters to LoRA, but conversion has two costs.
The first cost is accuracy; (IA)$^3$ and OFT change the base weight by a matrix of full rank, so a LoRA adapter of practical rank can only approximate them, and the quality of the approximation depends on what the adapter is trying to represent.
On adapters that we train, a rank-64 conversion of (IA)$^3$ keeps its accuracy on a mathematical reasoning task when it is fitted on the tenant's own data, but loses it when it is fitted on generic text.
The second cost is size.
A VeRA tenant owns only a few vectors, but as LoRA it needs its own factors, \hBytesSixtyFour\,GB at rank 64, leaving less GPU memory for other tenants and for the attention state of running requests.
An alternative to side-step these costs is to merge each adapter into its own copy of the model; this is exact but means requests for different adapters can no longer be run as a single batch on a shared base model.

To address this, we start from the kernel that serves LoRA, and understand how it can be adjusted to serve multiple tenants with different adapters.
In this LoRA kernel, for each layer, a batch holds tokens from many adapters that share one base weight $\bW$.
It then computes $\bh = \bx\bW$ once for the whole batch, and then computes each adapter's low-rank update in two small steps, a \emph{shrink} $\bx\bA$ and an \emph{expand} that multiplies the result by $\bB$ and adds it to $\bh$.
We generalize this structure with \emph{slots}, the places in this computation where an adapter can multiply or add a vector or matrix of its own.
These slots allow us to construct a kernel that can serve all the adapter shapes mentioned above.
Our contributions are as follows.

\textbf{A normal form and a fold rule} (\S\ref{sec:form}).
One formula, the \emph{normal form}, has six slots and covers LoRA, DoRA, (IA)$^3$, SSF, VeRA and OFT.
The \emph{fold rule} makes explicit what serving cost these slots induce.
A slot \emph{folds}, or merges into the adapter's own weights once at load time, if and only if the adapter owns a matrix or vector next to it.
The slots that do not fold stay \emph{live} and must be computed for every token.

\textbf{Three routes and their cost} (\S\ref{sec:cost}).
Live slots induce a cost and can be paid for in one of three ways, which we call \emph{routes}.
A byte model predicts the time that each route adds when its passes are limited by memory bandwidth.

\textbf{A compiler and a runtime} (\S\ref{sec:impl}).
A compiler turns a PEFT-library checkpoint of one of the five methods into the normal form, and a patch to vLLM, applied at run time, adds the routes to vLLM's LoRA path.
The resulting patched engine, Scute, can serve all five methods in one batch, with CUDA graphs and \texttt{torch.compile} on.
The engine, with tests and examples, is available at \url{https://github.com/ThomasWalker1/Scute-deploy}.

\textbf{Measurements} (\S\ref{sec:eval}).
Our engine matches the PEFT library with a mean error of at most \hAgreeWorst{} of each adapter's effect, and to within \hQualOursMaxDiff{} of its task accuracy.
On an RTX A6000, under a burst of requests that all arrive at once, a mix of four or five methods takes \hSDecRefLo--\hSDecRefHi{} more decode time and \hSPreRefLo--\hSPreRefHi{} more time to first token than the same number of LoRA adapters.
Under the same burst and with the same base GEMM in both engines (\S\ref{sec:eval:setup}), the best conversion that we tested, every tenant at rank 64, takes \hSVDecPct{} more decode time than our engine.
With 64 to 1024 (IA)$^3$ and VeRA tenants on the same GPU, our engine sustains \hCapSustainLo{} to at least \hCapSustainHi{} the request rate of their conversions within a latency target, with up to \hCapThousandThrTen{} their throughput.

\section{Background}
\label{sec:setup}

\textbf{Serving terms.}
A request first runs its whole prompt through the model in one step, called \emph{prefill}, and then produces one output token per step, called \emph{decode}.
We measure the time to first token (TTFT) and the time per output token after the first (TPOT).
An engine such as vLLM batches tokens from many requests in each step, and each running request needs memory for its attention state, the \emph{KV cache}.
An adapter is \emph{resident} when its weights are in GPU memory.
Other adapters wait in host memory, and the engine copies them in when a request needs them.
We also use \emph{tenant} to mean the tenant's adapter, and we call the set of adapters that an engine serves its \emph{pool}.

\textbf{GPU cost terms.}
A \emph{kernel} is a program that the host launches on the GPU, and each \emph{launch} has a fixed cost.
A \emph{pass} is a kernel that reads tensors from GPU memory and writes a result.
A GEMM is a general matrix multiplication.
Adapter computations perform few arithmetic operations per byte that they move, so their time is set by memory bandwidth, and we count their cost in bytes.
Unless stated otherwise, we measure on NVIDIA RTX A6000 GPUs (48\,GB, 6\,MiB L2 cache, PCIe) in fp16, and we repeat the main measurements on an NVIDIA A100-SXM4-80GB.
A CUDA graph records a sequence of launches once and replays it with one call from the host.
On the RTX A6000, a large copy within GPU memory reaches $\beta = \nBeta$\,GB/s, a copy that fits in the L2 cache reaches $\beta_{L2} = \nBetaLtwo$\,GB/s, and one launch inside a CUDA graph costs $\lambda = \nLambda\,\mu$s.

\textbf{The LoRA schedule.}
LoRA~\citep{lora} adds $c\,\bx\bA\bB$ to the base output $\bx\bW$.
Here $\bx \in \mathbb{R}^{N\times\dinn}$ holds one row per token, and $\bW \in \mathbb{R}^{\dinn\times\dout}$ is the base weight.
The adapter's factors $\bA \in \mathbb{R}^{\dinn\times r}$ and $\bB \in \mathbb{R}^{r\times\dout}$ have a small rank $r$, and $c$ is a scalar.
Punica's schedule~\citep{punica} sorts the batch so that the rows of each adapter are contiguous, and calls each such group of rows a \emph{segment}.
A vendor library GEMM computes $\bh = \bx\bW$ once for the whole batch.
The shrink then computes $\bV = \bx\bA \in \mathbb{R}^{N\times r}$ for each segment, and the expand reads $\bh$, adds $\bV\bB$, and writes the result.
A kernel splits its output into \emph{tiles}, blocks that one group of GPU threads computes, and in this schedule each tile of rows belongs to one segment, so a tile reads the factors of one adapter.
vLLM~\citep{pagedattention} fuses projections that share an input, such as the query, key and value projections $\bq,\bk,\bv$, into one layer with several output \emph{slices}, and its expand covers every slice in one launch.

\textbf{The methods.}
DoRA~\citep{dora} rescales each output column of $\bW + c\bA\bB$ to a learned magnitude $\bmm$.
(IA)$^3$~\citep{ia3} multiplies the keys, the values and the intermediate activation of the feed-forward network by learned vectors.
VeRA~\citep{vera} shares one frozen random pair $\bA,\bB$ of rank 256 among all adapters, and then each adapter learns two vectors per layer, $\bd$ inside the term and $\be$ after it.
OFT~\citep{oft} multiplies the input by a learned block-diagonal, orthogonal matrix $\bR$, with each block being computed from a skew-symmetric matrix using the Cayley transform.

\section{Extending the LoRA Schedule}
\label{sec:form}

Every method of \S\ref{sec:setup} changes the output of some of the model's linear projections, and it does so with a few small tensors of its own.
This section writes all of these changes in one formula, which extends LoRA's computation with a few extra parameters (\S\ref{sec:rule}).
It then shows which of these parameters can be merged into the adapter's own weights when the adapter loads, so that serving never computes them (\S\ref{sec:fold}).
Section~\ref{sec:cost} gives the cost of the parameters that remain.

\subsection{The Normal Form}
\label{sec:rule}

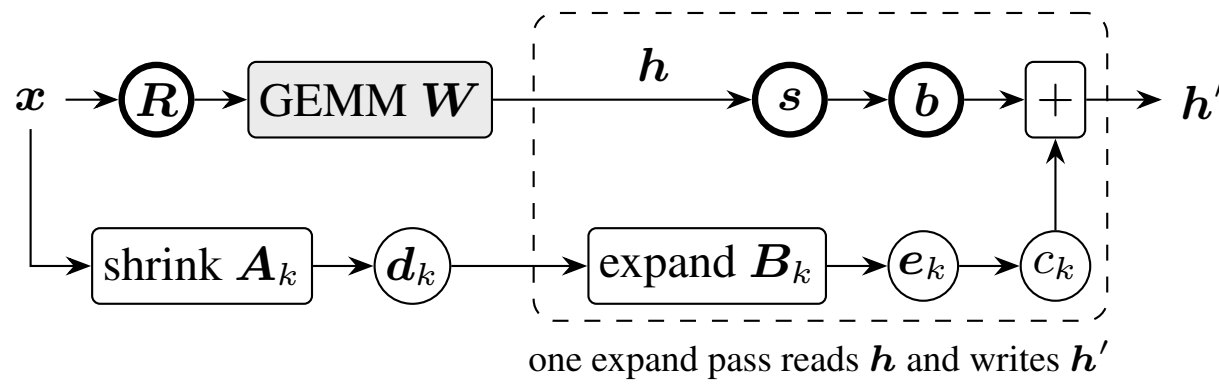
\begin{figure}[ht]
\centering
\begin{tikzpicture}[
  font=\footnotesize, >=Stealth, node distance=4mm,
  op/.style={draw, rounded corners=1.5pt, minimum height=5mm, inner sep=2pt},
  shared/.style={op, fill=black!8},
  slot/.style={circle, draw, inner sep=0.6pt, minimum size=4.6mm},
  live/.style={slot, very thick},
]
\node (x) {$\bx$};
\node[live, right=3.5mm of x] (R) {$\bR$};
\node[shared, right=3.5mm of R] (W) {GEMM $\bW$};
\node[op, anchor=west] (A) at ($(x.center)+(4mm,-11mm)$) {shrink $\bA_k$};
\node[slot, right=of A] (d) {$\bd_k$};
\node[op, right=9mm of d] (B) {expand $\bB_k$};
\node[slot, right=of B] (e) {$\be_k$};
\node[slot, right=of e] (c) {$c_k$};
\node[op] (sum) at (c |- W) {$+$};
\node[live, left=of sum] (b) {$\bb$};
\node[live, left=of b] (s) {$\bs$};
\node[right=5mm of sum] (y) {$\bh'$};
\draw[->] (x) -- (R); \draw[->] (R) -- (W); \draw[->] (W) -- node[above, pos=0.62] (hl) {$\bh$} (s);
\draw[->] (s) -- (b); \draw[->] (b) -- (sum); \draw[->] (sum) -- (y);
\draw[->] (x.south) |- (A.west); \draw[->] (A) -- (d); \draw[->] (d) -- (B); \draw[->] (B) -- (e);
\draw[->] (e) -- (c); \draw[->] (c) -- (sum);
\coordinate (Bl) at ($(B.west)+(-2.5mm,0)$);
\node[draw, dashed, rounded corners, inner xsep=3pt, inner ysep=3pt, fit=(hl)(s)(b)(sum)(B)(e)(c)(Bl),
      label={[font=\scriptsize]below:one expand pass reads $\bh$ and writes $\bh'$}] {};
\end{tikzpicture}
\caption{The LoRA schedule and its six slots. The shaded GEMM is shared by all adapters. Thick circles mark slots that never fold. Thin circles mark slots that fold when the adapter owns the factor next to them. The expand already reads $\bh$ and writes $\bh'$, so it can apply $\bs$, $\bb$ and $\be_k$ without an extra pass.}
\label{fig:schedule}
\end{figure}

An adapter changes some of the model's linear projections, such as the query projection of attention or the up projection of the feed-forward network, and leaves the others unchanged.
We call a projection that an adapter changes a \emph{hook}.
At each hook, the base model computes $\bx\bW$, and we write the output of adapter $a$ as
\begin{equation}
\begin{split}
F_a(\bx) = {} & \big((\bx\bR)\bW\big) \odot \bs + \bb \\
& + \textstyle\sum_{k=1}^{K} c_k \big( ( (\bx\bA_k) \odot \bd_k ) \bB_k \odot \be_k \big).
\end{split}
\label{eq:form}
\end{equation}
We call \eqref{eq:form} the \emph{normal form}.
Besides the low-rank factors $\bA_k$ and $\bB_k$, it has six \emph{slots}.
A slot is a parameter that an adapter can set, and that is the identity (a one, a vector of ones, a zero shift or the identity matrix) when the adapter does not use it.
There are three slots on the base path and three on each low-rank term.
On the base path, the input transform $\bR$ is a block-diagonal matrix with blocks of size $b_R$, the output scale $\bs \in \mathbb{R}^{\dout}$ multiplies each output column, and the output shift $\bb \in \mathbb{R}^{\dout}$ is added to it.
Each term $k$, of rank $r_k$, has an inner vector $\bd_k \in \mathbb{R}^{r_k}$, an outer vector $\be_k \in \mathbb{R}^{\dout}$ and a scalar $c_k$.
A vector multiplies each row of the tensor on its left, element by element.
Each tensor is either \emph{owned}, which means that each adapter stores its own copy, or \emph{shared} by all adapters in the pool, and the base weight $\bW$ is always shared.
A method is a \emph{member} of the normal form if \eqref{eq:form} can express it.
Figure~\ref{fig:schedule} shows where each slot sits in the schedule.

Table~\ref{tab:members} shows how each method fits into the normal form.
For example, (IA)$^3$ scales the intermediate activation of a gated feed-forward network.
With gate weight $\bW_g$, up weight $\bW_u$ and activation function $\sigma$, this network computes $\sigma(\bx\bW_g)\odot\bx\bW_u$.
Since $(\sigma(\bx\bW_g)\odot\bx\bW_u)\odot\bl = \sigma(\bx\bW_g)\odot(\bx\bW_u\odot\bl)$, the vector $\bl$ of (IA)$^3$ is the slot $\bs$ of the up projection.
Similarly, OFT, as the PEFT library implements it, rotates $\bx$ and then calls the unchanged base layer, so its rotation is the slot $\bR$.

Other methods fit as well.
LoHa~\citep{fedpara} and LoKr~\citep{krona} build the update from products of small factors, and they are members once these factors are expanded at load into a low-rank term, whose rank can be larger than the adapter's own rank.
BOFT~\citep{boft} is a variant of OFT whose rotation is a product of several permuted block-diagonal factors, so it needs one input transform for each factor, an extension of the single $\bR$ that we do not implement.
SSF is a member, but the PEFT library (version 0.20) has no SSF method, so we do not consider it.
HiRA~\citep{hira} and full fine-tuning are not members, because they need a dense matrix for each adapter after the shared GEMM.

\begin{table}[ht]
\centering
\caption{Members of the normal form \eqref{eq:form}; we identify which matrices each adapter owns (VeRA's $\bA,\bB$ are shared), the folds that Proposition~\ref{prop:fold} allows, and the slots that stay live.}
\label{tab:members}
\small
\vspace{2pt}
\begin{tabular}{llll}
\toprule
method & owned matrices & folds & live slots \\
\midrule
LoRA & $\bA,\bB$ & $c \to \bA$ & none \\
DoRA & $\bA,\bB$ & $\be \to \bB$, $c \to \bA$ & $\bs$ \\
(IA)$^3$ & none & none & $\bs$ \\
SSF & none & none & $\bs$, $\bb$ \\
VeRA & none & $c \to \bd$ & $\bd$, $\be$ \\
OFT & $\bR$ & none & $\bR$ \\
\bottomrule
\end{tabular}
\end{table}

\subsection{The Fold Rule}
\label{sec:fold}

If a slot multiplies a matrix that belongs to the adapter, the engine can just multiply the two once, when the adapter loads.
Thus, the slot disappears from serving.
For example, $(\bx\bA)\odot\bd = \bx(\bA\,\mathrm{diag}(\bd))$, so an inner vector $\bd$ folds into an owned $\bA$.
We say that such a slot \emph{folds}, and we call a slot that does not fold \emph{live}.
A fold may change only tensors that the adapter owns, so it never changes the shared weight $\bW$ or factors that other adapters use (App.~\ref{app:proofs} gives the exact definition).

\begin{proposition}[Fold rule, informal]
\label{prop:fold}
For adapters of low total rank and generic parameters, a slot folds if and only if the adapter owns a matrix or vector next to it in the product that applies it.
In particular, the output scale $\bs$, the output shift $\bb$ and the input transform $\bR$ never fold, because only the shared $\bW$ is next to them.
\end{proposition}

Proposition~\ref{propA:fold} in the appendix states the exact conditions, which are satisfied by every member in Table~\ref{tab:members}.
The identities $(\bx\bA)\odot\bd = \bx(\bA\,\mathrm{diag}\,\bd)$, $((\bx\bA)\odot\bd)\bB = (\bx\bA)(\mathrm{diag}(\bd)\bB)$ and $(\bV\bB)\odot\be = \bV(\bB\,\mathrm{diag}\,\be)$ prove the ``if'' direction.
The ``only if'' direction is a rank argument, since a slot next to shared tensors changes the output by a matrix of high rank that the adapter's owned low-rank factors cannot absorb.
For the output scale $\bs$, the same argument shows what an exact fold would need.

\begin{corollary}[Folding $\bs$, informal]
\label{cor:s}
To remove an output scale $\bs$ exactly, a member needs low-rank terms of total rank at least $\mathrm{rank}(\bW\,\mathrm{diag}(\bs-\mathbf{1}))$ minus the adapter's own rank (Corollary~\ref{corA:s}).
\end{corollary}

For a trained DoRA adapter, the scale $\bs$ is close to one, with a root-mean-square deviation of \nFoldGdev{}, but it differs from one in every column.
Yet $\bW\,\mathrm{diag}(\bs-\mathbf{1})$ has full numerical rank on \nFoldFullRank{} adapted projections of Llama-3.1-8B, and even the best rank-16 approximation of it leaves at least \nFoldLeftSixteen{} of its squared Frobenius norm (App.~\ref{app:foldrank}).
\textbf{Hence, to fold DoRA's scale exactly, an engine must merge the adapter into a copy of the base weight.}
This means that each DoRA tenant would then need its own copy of $\bW$ and its own GEMM, so its requests could no longer share a batch with other tenants.
A batched engine must therefore keep $\bs$ live and apply it at serving time.
In \S\ref{sec:cost}, we show that the expand can do this at almost no extra cost.

VeRA's scalar folds into its own vectors, but the vectors $\bd$ and $\be$ stay live, because the factors next to them in \eqref{eq:form} are shared.
Thus, in contrast to DoRA, keeping the slots live is what makes VeRA cheap to serve.
Folding $\bd$ and $\be$ would otherwise give each tenant its own copy of $\bA$ and $\bB$, which is equivalent to a conversion to a rank-256 LoRA adapter.
With the slots live, all VeRA tenants share one copy of the factors, which a batch reads once per step (\S\ref{sec:cost}).

OFT's rotation sits before the shared $\bW$ and cannot fold, but it still keeps the GEMM shared, because $(\bx\bR_a)\bW$ for every adapter $a$ in a batch is one GEMM over the transformed rows.

\textbf{Stacking.}
A request may also apply two adapters at one hook, one after the other.
For example, a tenant may add its own LoRA adapter on top of an (IA)$^3$ adapter.
Applying a member $F_2$ of \eqref{eq:form} to the output of another member $F_1$ of \eqref{eq:form} gives again a member of \eqref{eq:form} (App.~\ref{app:proofs}).
Its output scale is $\bs_1\odot\bs_2$ and its shift is $\bb_1\odot\bs_2+\bb_2$.
It holds the terms of both members, where the scale $\bs_2$ of the second member also multiplies each term of the first, through its outer vector $\be$.
The stack can therefore run as one segment with more terms, without an extra launch.
However, it is important to note that the order of the two members matters.
In the example, (IA)$^3$ after LoRA scales the LoRA update by $\bs_2$, while LoRA after (IA)$^3$ adds the update unscaled.
In general, the two orders differ when one member has an output scale $\bs \neq \mathbf{1}$ and the other has a shift or a term.

\section{The Cost of Serving Methods Beyond LoRA}
\label{sec:cost}

Section~\ref{sec:form} showed which slots fold into an adapter's own weights.
A folded slot costs nothing at serving time, but a live slot must be computed for every token.
This section asks how a LoRA schedule can compute each live slot, and what that costs.
We first describe three ways to compute a live slot, which we call routes, and count the bytes that each one moves (\S\ref{sec:routes}).
We then count the bytes of the alternative, converting each adapter to LoRA (\S\ref{sec:conv}).
Finally, we turn these counts into a model of time and test it on the GPU (\S\ref{sec:bytemodel}).

\subsection{The Cost of Serving Through the Normal Form}
\label{sec:routes}

A live slot must be computed inside the LoRA schedule of \S\ref{sec:setup}, which runs the shared GEMM, the shrink and the expand for each layer.
We call a way to do this a \emph{route}.
We introduce three routes to compute the live slots of the five members we are considering.

First, the output scale $\bs$ of DoRA and (IA)$^3$ multiplies $\bh$, the output of the shared GEMM. The first route applies $\bs$ inside the expand. Second, the vectors $\bd$ and $\be$ of VeRA sit next to factors that all VeRA tenants share. The second route computes this term with two GEMMs over the whole batch. Third, the rotation $\bR$ of OFT multiplies the input $\bx$ before the shared GEMM. The third route applies $\bR$ in a pass of its own.

Since adapter passes are limited by memory bandwidth (\S\ref{sec:setup}), the cost of a route can be quantified by the number of bytes that its passes read and write.
Proposition~\ref{prop:cost} provides an overview of these bytes, with Propositions~\ref{propA:out}--\ref{propA:shared} in the appendix giving the exact counts.

\begin{proposition}[Cost of a live slot, informal]
\label{prop:cost}
Consider a batch of $N$ tokens from $k$ adapters.
\begin{enumerate}
\setlength{\itemsep}{0pt}
\item[(i)] A slot carried by a pass that already reads and writes its tensor adds one read of each adapter's vector, at most $k\dout$ elements.
\item[(ii)] A slot on the output in a pass of its own adds $2N\dout$ elements and one launch, and $4N\dout$ elements if the pass first writes a copy of the slot for every row.
\item[(iii)] An input transform in a pass of its own adds $2N\dinn$ elements plus its blocks, and it does too little arithmetic per byte to be limited by anything but bandwidth.
\item[(iv)] A term on factors that all $k$ adapters share reads the factors once per step, while their LoRA conversions read them $k$ times.
\end{enumerate}
\end{proposition}

\textbf{Route one applies the scale $\bs$ in the expand.}
The expand reads $\bh$, adds each adapter's low-rank update, and writes the result, one tile of rows at a time.
Each tile belongs to one adapter, so the expand can also multiply $\bh$ by that adapter's $\bs$ before it writes the tile.
This adds one read of $\bs$ per tile, at most $k\dout$ elements per step (Proposition~\ref{prop:cost}(i)), which is small next to the $2N\dout$ elements that the expand already moves.
The alternative is a pass of its own after the expand, which reads and writes the output a second time.
By Proposition~\ref{prop:cost}(ii), this pass doubles the expand's traffic, and it triples that traffic if it first writes a copy of $\bs$ for every row.

\textbf{Route two computes shared factors in GEMMs over row tiles.}
VeRA's vectors $\bd_a$ and $\be_a$ belong to each adapter $a$, but the factors $\bA$ and $\bB$ next to them are shared, so the vectors stay live.
For this term, the cost depends on the order in which a kernel visits the batch.
The segmented expand visits the rows of one adapter at a time, so it reads $\bB$ once for each adapter, as $k$ conversions to LoRA would.
A GEMM that splits the batch into tiles of rows, whatever their adapters, reads each tile of $\bB$ once for all the rows of a tile.
The second route therefore runs two such GEMMs after the expand.
The first computes $\bx\bA$ for every row of the batch and multiplies each row by its adapter's $\bd_a$.
The second multiplies the result by $\bB$, multiplies each row by its adapter's $\be_a$ as each tile finishes, and adds the result to the output, skipping rows whose adapter has no such term.
The factors are then read once per step, however many VeRA tenants the batch holds (Proposition~\ref{prop:cost}(iv)), and the vectors add only $k(r+\dout)$ elements.

\textbf{Route three applies the rotation in a pass of its own.}
OFT's rotation $\bR$ multiplies the input before the shared GEMM, and the expand runs after that GEMM, so the expand cannot carry $\bR$.
The third route applies $\bR$ in a pass of its own before the shared GEMM.
This pass reads each row of the input, multiplies it by the blocks of that row's adapter, and writes the result, which adds $2N\dinn$ elements plus the blocks (Proposition~\ref{prop:cost}(iii)).
Because $\bR$ is block diagonal, each output element needs only one small block, so the pass does little arithmetic per byte and stays limited by bandwidth.
The rotated rows of all adapters then go through the shared GEMM, so OFT tenants still share $\bW$ with every other tenant.

\subsection{The Cost of Serving Through Conversion}
\label{sec:conv}

The alternative to a live slot is to convert the adapter to LoRA and serve with the existing kernel.
The update of (IA)$^3$, $\bW\,\mathrm{diag}(\bs-\mathbf{1})$, has full rank whenever $\bW$ has full rank and every entry of $\bs$ differs from one, and the update of OFT, $(\bR-\bI)\bW$, has full rank for a generic $\bR$.
A rank-$r$ LoRA conversion therefore only approximates them, and it reads $r(\dinn+\dout)$ elements per adapter per step.
Table~\ref{tab:routes} gives these bytes for Llama-3.1-8B and a batch of 16 tenants, and for each method the route reads fewer bytes than the conversions.
VeRA's shared factors are the one large read, and that read happens once per step, however many VeRA tenants the batch holds.
\textbf{A route keeps each adapter exact and reads less than its conversion, and VeRA's shared factors cost less than one exact conversion and less than rank-64 conversions from four tenants.}

\begin{table}[ht]
\centering
\caption{The bytes that each route reads per decode step with 16 tenants in the batch, for Llama-3.1-8B (fp16, every adapted projection of 32 layers), and that time at copy bandwidth. \ding{51}: the route serves the adapter exactly. Bold: the fewest bytes for each method.}
\label{tab:routes}
\small
\setlength{\tabcolsep}{2.5pt}
\begin{tabular}{llcrr}
\toprule
method & route & exact & GB & ms \\
\midrule
LoRA, rank 16 & expand & \ding{51} & 1.34 & 1.97 \\
VeRA, rank 256 & route 2 (shared factors) & \ding{51} & \textbf{1.19} & \textbf{1.75} \\
 & as LoRA, rank 256 & \ding{51} & 21.47 & 31.58 \\
 & as LoRA, rank 64 & \ding{55} & 5.37 & 7.90 \\
OFT, blocks of 32 & route 3 (pre-transform) & \ding{51} & \textbf{1.28} & \textbf{1.88} \\
 & as LoRA, rank 64 & \ding{55} & 5.37 & 7.90 \\
(IA)$^3$ & route 1 (expand) & \ding{51} & \textbf{0.01} & \textbf{0.01} \\
 & as LoRA, rank 64 & \ding{55} & 1.88 & 2.76 \\
\bottomrule
\end{tabular}

\end{table}

\subsection{A Byte Model for Predicting Time Cost}
\label{sec:bytemodel}

We predict a route's time as $t_0 + \Delta Q/\beta + \Delta L\,\lambda$.
Here $t_0$ is the measured time without the extra passes, and $\Delta Q$ and $\Delta L$ are the bytes and launches that Proposition~\ref{prop:cost} adds.
When the extra read and write fit in the L2 cache together, $\beta$ becomes $\beta_{L2}$.

\begin{figure}[ht]
\centering
\includegraphics[width=\columnwidth]{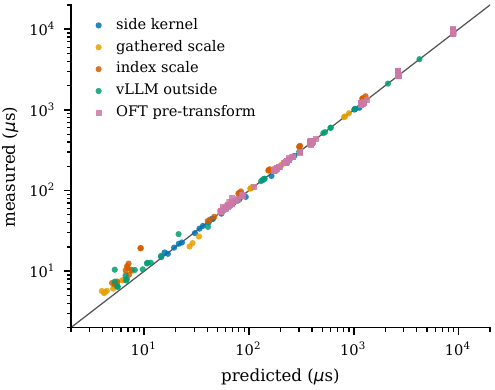}
\caption{Predicted against measured time on the RTX A6000, for OFT's input transform and for four routes that apply DoRA's scale outside the expand (App.~\ref{app:bytemodel}). The line marks equality.}
\label{fig:bytemodel}
\end{figure}

We test this model on OFT's input transform and on four routes that apply DoRA's $\bs$ outside the expand (App.~\ref{app:bytemodel}).
Figure~\ref{fig:bytemodel} shows every measured cell, and where the extra bytes dominate the launch cost, the median error on total time is \nBmErrOutside--\nBmErrIndex{} per route.
\textbf{Where bytes dominate, the byte model predicts the time of a route from its bytes alone.}

\textbf{DoRA in vLLM.}
The byte counts also explain what DoRA costs in a real engine, where the choice is between the first route and a pass of its own after the expand.
We measure both with vLLM's own kernels, over every LoRA layer of Llama-3.1-8B.
With DoRA's $\bs$ carried in the expand (route one), the expand takes \nEngSlotLo--\nEngSlotHi{} the time of vLLM's LoRA expand for 1024 tokens or more, as Proposition~\ref{prop:cost}(i) predicts.
An open vLLM pull request instead writes a copy of the scale for every row and multiplies the output in a pass after the expand.
This costs \nEngOutLo--\nEngOutHi{} the time of the LoRA expand, close to the $3\times$ that Proposition~\ref{prop:cost}(ii) gives for such a pass.
End to end, under a decode burst, four DoRA tenants served through our slot take \nMeasSlotDecRatio{} the time per output token of four LoRA tenants (App.~\ref{app:measure}).
\textbf{Carried in the expand, DoRA's scale costs about as much as plain LoRA.}

\section{Scute}
\label{sec:impl}

We build these routes into vLLM, a production serving engine, without forking it.
The system has two parts.
An admission compiler turns each adapter into the normal form when the engine loads it (\S\ref{sec:compiler}), and a runtime patch adds the routes to vLLM's LoRA path (\S\ref{sec:runtime}).

\subsection{The Admission Compiler}
\label{sec:compiler}

Our compiler reads a checkpoint of the PEFT library and writes the adapter in normal form under vLLM's LoRA tensor names, so that vLLM's own loader, paging and scheduler handle it.
LoRA and DoRA pass through unchanged.
When the engine admits a DoRA adapter, it computes the slot $\bs = \bmm\oslash\lVert\bW + c\bA\bB\rVert_c$ from the magnitude $\bmm$, where $\oslash$ is element-wise division and $\lVert\cdot\rVert_c$ is the norm of each output column.
The engine computes $\bs$ at admission because it needs the base weight in the engine's own layout.
(IA)$^3$ becomes the slot $\bs$, on the up slice for the feed-forward network (\S\ref{sec:rule}).
VeRA becomes the vectors $\bd$ and $\be$ plus a reference to the shared pair, which is stored once.
OFT's parameters become explicit blocks $\bR$, computed in the parametrization that the adapter was trained with.
vLLM's loader requires a term, so a method without one gets a zero rank-one term, which the expand skips.

\subsection{The Runtime}
\label{sec:runtime}

The integration patches vLLM 0.19.1 at run time and leaves the scheduler, paging, attention and batch composition unchanged (App.~\ref{app:engine}).
Each adapted layer gets one buffer per slot, $\bs$, $\bd$, $\be$ and $\bR$, allocated with the LoRA weights and set to the identity.
The layer's forward pass runs four steps.
First, the base GEMM runs, split by slice when an adapter has an $\bR$, because $\bq$, $\bk$ and $\bv$ each have their own rotation.
Second, vLLM's shrink runs, and it skips adapters without a term on the layer.
Third, vLLM's expand runs with $\bs$ applied as each tile finishes, which is a one-line change to the kernel, \texttt{tiled\_c += tiled\_out * s}.
Fourth, the shared term of Proposition~\ref{prop:cost}(iv) runs.

The input transform and the shared term are Triton~\citep{triton} kernels that run every slice of a layer in one launch.
We register them as custom PyTorch operators, which \texttt{torch.compile} treats as black boxes and CUDA graphs capture.
Outside captured CUDA graphs, a step whose batch holds no OFT or VeRA adapter skips the split GEMM and the shared term.
Tensor parallelism splits each layer across several GPUs.
Each GPU then reads its own part of each factor, vector and block, and the layer's existing all-reduce, a sum across GPUs, adds the partial terms.

\begin{figure*}[ht]
\centering
\includegraphics[width=0.82\textwidth]{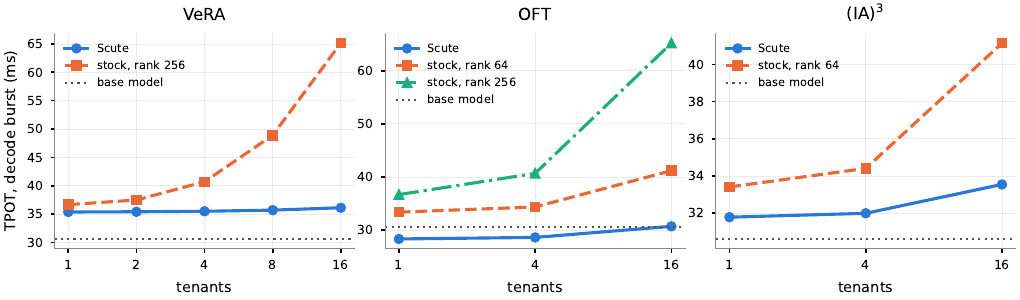}
\caption{Time per output token under a decode burst against the number of tenants of one method, RTX A6000. Scute serves each adapter exactly. Stock serves each adapter converted to LoRA, at rank 64 and, for OFT, also at rank 256, with every adapter resident. The base model, without adapters, is shown for reference. Stock and the base model run as shipped. With OFT, Scute runs the split base GEMM of \S\ref{sec:eval:setup}, which is faster than the fused base GEMM. Median of six repetitions.}
\label{fig:scaling}
\end{figure*}

\section{Evaluating Scute}
\label{sec:eval}

We now perform evaluations to answer the following.
First, does Scute compute each method exactly, and how much accuracy would a conversion to LoRA lose (\S\ref{sec:eval:quality})?
Second, how does the cost grow with the number of tenants of one method (\S\ref{sec:eval:scaling})?
Third, what does a pool of several methods cost, against LoRA-only and against the conversions (\S\ref{sec:eval:mixed})?
Fourth, what does the small size of the normal form give when many tenants share one GPU (\S\ref{sec:eval:capacity})?
Section~\ref{sec:eval:a100} repeats the main results on a second GPU.

\subsection{Setup}
\label{sec:eval:setup}

We use vLLM 0.19.1 and Llama-3.1-8B in fp16, with CUDA graphs and \texttt{torch.compile} on.
vLLM's LoRA kernels are tuned for the GPU at every rank that an engine serves.
We train fifteen adapters with the PEFT library and its default settings, one for each of LoRA, DoRA, (IA)$^3$, VeRA and OFT on each of GSM8K, MNLI and BoolQ (App.~\ref{app:training}).
Serving cost depends only on an adapter's shapes, so we measure it on synthetic tenants with the shapes of the trained adapters.

We compare four engines.
\emph{Ours} serves every adapter in normal form.
\emph{Stock} is vLLM serving every adapter converted to LoRA, VeRA at rank 256 and the other methods at rank 64.
\emph{LoRA-only} is vLLM serving the same number of rank-16 LoRA tenants, which shows the price of the other methods.
The \emph{base model} runs without adapters.
vLLM runs every adapter at the largest rank that the engine serves, and pads smaller adapters with zeros.

Our slot buffers sit outside vLLM's memory budget. We therefore give our engine 0.83 of device memory against 0.90 for the others, so that both use the same device memory, \hFMMemLo--\hFMMemHi\,GiB in the 20-tenant pool.
Stock keeps every adapter resident, except in the 20-tenant pool with VeRA at rank 256, where 16 of the 20 fit.

A prefill-heavy load sends 256-token prompts for 8 output tokens, and a decode-heavy load sends 8-token prompts for 128 output tokens.
Requests arrive at a fixed rate as a Poisson process, or all at once, which we call a \emph{burst}.
Tenants take turns, or, in one pool, follow a Zipf law, under which a few tenants receive most requests.
Each measurement runs six repetitions with new prompts and arrival times, and each measurement with many tenants (\S\ref{sec:eval:capacity}) runs three.
The engines of a repetition share these and run in every order equally often (App.~\ref{app:measure}).
We report the median of the per-repetition ratios, comparing median TPOT for decode-heavy load and mean TTFT for prefill-heavy load.

When an engine serves OFT, our runtime computes each fused base GEMM as one GEMM per slice (\S\ref{sec:impl}).
On the RTX A6000, this split alone makes the base model faster, \hSBaseSplit{} against \hSBaseFused\,ms per output token.
For the pools that hold OFT, we therefore also run stock, LoRA-only and the base model with the same split.
We call this the \emph{same GEMM path}, and we call vLLM without the split \emph{as shipped}.

\subsection{Exactness and Accuracy}
\label{sec:eval:quality}

We first check that our engine computes what the PEFT library computes.
Trained vectors are close to the identity, so a slot that is dropped, sharded wrongly or applied twice rarely changes the greedy tokens, and we therefore use \emph{amplified} adapters, whose vectors move the log-probabilities by a clear margin.
We serve one batch with \hAgreeAdapters{} amplified adapters of all five methods and compare log-probabilities with the PEFT library in float32.
For each method, the mean error is \hAgreeBest--\hAgreeWorst{} of the adapter's effect, which is its change to the log-probabilities of the base model (App.~\ref{app:engine}).
The largest errors belong to the smallest effects, and they match the fp16 error of a plain LoRA adapter, and the same holds on two GPUs with tensor parallelism.
The fifteen trained adapters, served together, keep every task metric within \hQualOursMaxDiff{} of the PEFT library (Tables~\ref{tab:quality} and~\ref{tab:qualityfull}).
Here and below, a point is one percentage point of accuracy.
\textbf{Our engine serves all five methods in one batch and matches the PEFT library with a mean error of at most \hAgreeWorst{} of each adapter's effect, and to within \hQualOursMaxDiff{} of its accuracy.}

\begin{table}[ht]
\centering
\caption{Task accuracy of the trained (IA)$^3$, OFT and VeRA adapters. \emph{PEFT} is the accuracy in percent when the PEFT library runs the adapter. The other columns give the difference from PEFT in points, in Scute (all fifteen adapters in one batch) and converted to a LoRA of rank 64. Bold marks the smallest absolute difference in each row. SVD is a truncated SVD of the update. ASVD is activation-aware SVD fitted on 128 examples of the tenant's own task, and ASVD$_g$ is the same fitted on generic text. Dist.\ is distillation on the tenant's task. VeRA was not distilled. App.~\ref{app:conversions} gives other ranks and fits.}
\label{tab:quality}
\small
\setlength{\tabcolsep}{2.4pt}
\begin{tabular}{llrrrrrr}
\toprule
 & & & & \multicolumn{4}{c}{LoRA of rank 64} \\
\cmidrule(lr){5-8}
method & task & PEFT & Scute & SVD & ASVD & ASVD$_g$ & dist. \\
\midrule
(IA)$^3$ & GSM8K & 59.8 & \textbf{+0.4} & $-$18.6 & $-$0.8 & $-$19.6 & \textbf{{\boldmath$-$}0.4} \\
 & MNLI & 90.2 & \textbf{0.0} & $-$23.2 & +0.5 & $-$16.9 & $-$0.3 \\
 & BoolQ & 87.9 & \textbf{{\boldmath$-$}0.1} & $-$4.8 & +0.8 & $-$1.6 & +0.3 \\
\addlinespace
OFT & GSM8K & 63.0 & \textbf{+0.2} & $-$5.6 & $-$1.2 & $-$4.6 & $-$2.0 \\
 & MNLI & 90.9 & \textbf{0.0} & $-$2.9 & +0.3 & $-$0.9 & $-$0.1 \\
 & BoolQ & 89.7 & \textbf{0.0} & $-$0.6 & $-$0.1 & $-$0.2 & $-$0.8 \\
\addlinespace
VeRA & GSM8K & 58.4 & \textbf{+0.2} & +0.8 & $-$0.8 & +3.4 & --- \\
 & MNLI & 90.9 & +0.1 & $-$0.6 & +0.3 & \textbf{0.0} & --- \\
 & BoolQ & 89.5 & \textbf{0.0} & $-$0.1 & +0.1 & $-$0.1 & --- \\
\bottomrule
\end{tabular}

\end{table}

Next, we convert the six (IA)$^3$ and OFT adapters at ranks 16 and 64.
Truncated SVD, which minimizes the error in the weights, falls more than 3 points below the PEFT library in \hConvSvdBelow{} of the 12 cases.
Activation-aware SVD~\citep{asvd} minimizes the error on sample inputs, and when it is fitted on the tenant's own task data it stays within \hConvAsvdOwnWorstSixtyFour{} points of the PEFT library at rank 64 on all six adapters.
Distillation on the same data stays within \hConvDistillWorstSixtyFour{} points.
Fitted on generic text, activation-aware SVD falls more than 3 points below the PEFT library in \hConvAsvdGenBelow{} of the 12 cases, and (IA)$^3$ loses up to \hConvAsvdGenWorstSixtyFour{} points at rank 64.
VeRA converts well by plain SVD at rank 64, within \hConvVeraSvdWorstSixtyFour{} points, but it loses up to \hConvVeraSvdWorstSixteen{} points at rank 16.
DoRA converts within noise, because its magnitude changes little (App.~\ref{app:training}).
\textbf{Converted to a rank-64 LoRA without access to the original task data, (IA)$^3$ loses up to \hConvSvdWorstSixtyFour{} points of accuracy, and a rank-16 conversion of VeRA loses up to \hConvVeraSvdWorstSixteen{} points. Our engine serves every adapter exactly from its checkpoint alone.}

\subsection{Cost per Tenant}
\label{sec:eval:scaling}

Next, we isolate what each tenant costs.
Figure~\ref{fig:scaling} grows the number of tenants of one method, with every adapter resident in both engines.

Our engine reads VeRA's shared factors once per step, so its time stays flat, \hFVOursOne\,ms per token at one tenant and \hFVOursSixteen\,ms at 16.
Converted, each VeRA tenant brings its own rank-256 factors, so stock takes \hFVStockOne\,ms at one tenant and \hFVStockSixteen\,ms at 16, which is \hFVSixteen{} our time.

OFT's blocks are as small as the factors of a rank-16 LoRA, and (IA)$^3$ owns only vectors.
At 16 tenants, the rank-64 conversion of (IA)$^3$ takes \hFISixteen{} our time, and that of OFT takes \hSOSixteen{} with the same GEMM path (\hSOSixteenShipped{} as shipped).
Each conversion adds decode time per tenant within 10\% of the byte model's prediction, while our engine adds \hFVOursSlope--\hFOOursSlope\,ms per tenant (App.~\ref{app:measure}).
Under a prefill burst, stock takes \hFPreOneLo--\hFPreOneHi{} our mean TTFT at one tenant and \hFPreSixteenLo--\hFPreSixteenHi{} at 16.

\textbf{Each tenant converted to LoRA adds the time to read its own factors, as the byte model predicts, while each tenant in our engine adds only \hFVOursSlope--\hFOOursSlope\,ms per token.}

\begin{figure*}[t]
\centering
\includegraphics[width=0.82\textwidth]{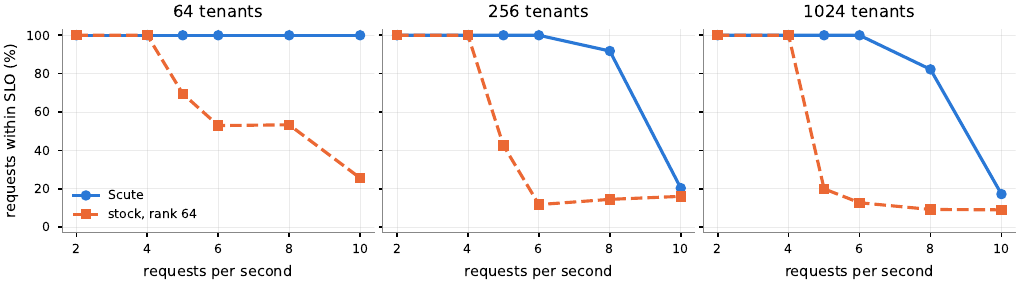}
\caption{Share of requests that meet the latency target (first token within 1\,s, then at most 100\,ms per token) against arrival rate, under decode-heavy load on one RTX A6000. The pools hold 64, 256 and 1024 tenants, half (IA)$^3$ and half VeRA. Scute serves them exactly, and stock serves their rank-64 conversions. Median of three repetitions.}
\label{fig:capacity}
\end{figure*}

\subsection{Pools of Methods}
\label{sec:eval:mixed}

\begin{table}[ht]
\centering
\caption{Here we measure serving times for pools of several methods in one engine, under a burst, RTX A6000. The \emph{five methods} pool holds 20 tenants, four of each method, and stock keeps 16 of them resident. \emph{VeRA at 64} is a separate run of the same pool, with stock serving VeRA converted at rank 64 instead of 256. The \emph{no VeRA} pool holds 16 tenants, four each of LoRA, DoRA, (IA)$^3$ and OFT. We measure the ratios of serving times for Scute against serving the same number of plain rank-16 LoRA adapters, for the conversions against Scute on the same GEMM path, and for the conversions on vLLM as shipped against Scute. Ratios are the median over six repetitions. App.~\ref{app:measure} gives every rate of vLLM as shipped.}
\label{tab:mixed}
\small
\setlength{\tabcolsep}{2.8pt}
\begin{tabular}{llccc}
\toprule
 & & Scute / & Conversion / & Conversion / \\
pool & load & LoRA-only & Scute & Scute (stock) \\
\midrule
five methods & decode & 1.17 & 1.65 & 1.73 \\
 & prefill & 1.18 & 1.14 & 1.12 \\
VeRA at 64 & decode & 1.11 & 1.12 & --- \\
 & prefill & 1.12 & 0.94 & --- \\
no VeRA & decode & 1.05 & 1.14 & 1.25 \\
 & prefill & 1.04 & 1.01 & 0.99 \\
\bottomrule
\end{tabular}

\end{table}

\begin{figure}[ht]
\centering
\includegraphics[width=0.92\columnwidth]{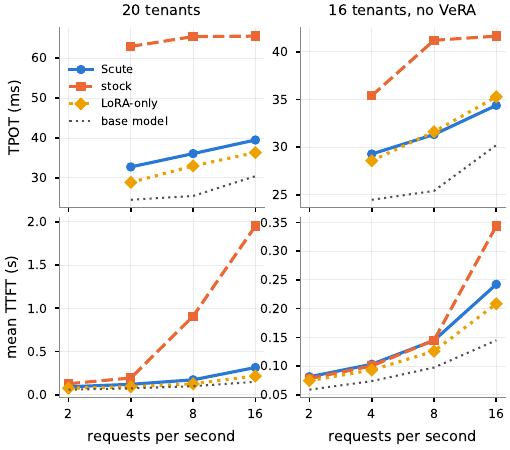}
\caption{Latency against arrival rate in the two pools, RTX A6000. Scute decodes faster than stock at every rate, and under prefill-heavy load stock's TTFT grows fastest. Stock, LoRA-only and the base model run as shipped, so the gap between Scute and LoRA-only differs from the same-GEMM-path ratio of Table~\ref{tab:mixed}. Top, median TPOT under decode-heavy load. Bottom, mean TTFT under prefill-heavy load. Median of six repetitions.}
\label{fig:loadlat}
\end{figure}

A platform serves several methods at once, so we now mix them in one engine.
Table~\ref{tab:mixed} gives the controlled comparison, under a burst, with every engine on the same GEMM path.
Figure~\ref{fig:loadlat} then shows that the result holds at every arrival rate in vLLM as shipped.

\textbf{The price of exactness.}
We compare our engine with LoRA-only, both on the same GEMM path.
Under a burst, the pool of all five methods takes \hSMDecRef{} the decode time and \hSMPreRef{} the TTFT of LoRA-only, and the pool of four methods without VeRA takes \hSMnvDecRef{} and \hSMnvPreRef.
The extra time comes from the routes, which are VeRA's shared term, OFT's input transform, and the vectors that each step reads.

\textbf{Against the conversions.}
VeRA's conversion is exact at rank 256 and keeps its accuracy at rank 64 (\S\ref{sec:eval:quality}).
With the rank-64 conversion, stock pads every adapter only to rank 64 and keeps all 20 tenants resident, yet on the same GEMM path it still takes \hSVDec{} our decode time under a burst and \hSVDecSixteen{} at 16 requests per second, because each tenant brings its own factors.
With VeRA at rank 256, the conversions take \hSMDec{} our decode time.

\textbf{At every rate.}
Figure~\ref{fig:loadlat} shows the two main pools from 2 to 16 requests per second.
vLLM as shipped takes \hShippedDecLo--\hShippedDecHi{} our decode time in every pool and at every rate, including under Zipf popularity and with two GPUs (App.~\ref{app:measure}).
In the five-method pool, most of this time comes from padding.
vLLM runs every adapter at the largest rank in the engine, here 256, and it also runs zero factors on the projections that a tenant does not adapt.
The bytes of the conversions at their own ranks would take \hFDecMOwn\,ms per step, and padding raises this to \hFDecMPad\,ms, close to the measured \hFAddMStock\,ms (App.~\ref{app:measure}).
Under prefill-heavy load, stock's TTFT grows fastest once the engine saturates, to \hFMPreHi{} ours at 16 requests per second.

\textbf{Under a burst on one RTX A6000, exact serving of a mix of methods takes only \hSDecRefLo--\hSDecRefHi{} more decode time than serving LoRA adapters alone. Converted to LoRA, each mix takes at least \hSVDecPct{} more decode time than our engine on the same GEMM path, and vLLM as shipped takes \hShippedDecLo--\hShippedDecHi{} our decode time at every rate.}

\subsection{Many Tenants}
\label{sec:eval:capacity}

The normal form keeps each adapter small.
An (IA)$^3$ or VeRA tenant owns a few vectors, a few megabytes in total, and all VeRA tenants share one pair of factors, while the rank-64 LoRA that keeps VeRA's accuracy is \hBytesSixtyFour\,GB.
On one RTX A6000, our engine can hold all \hCapOursResident{} tenants resident with a KV cache of \hCapOursResidentKv{} tokens, while stock can hold \hCapStockResidentMax{} rank-64 conversions, with a cache of \hCapStockResidentKv{} tokens.

We serve 64, 256 and 1024 tenants, half (IA)$^3$ and half VeRA, in turn under decode-heavy load.
We limit our engine to 64 resident tenants, close to the 48 of stock, so that both engines copy adapters into GPU memory when a request needs them.
Our engine keeps the other tenants in host memory, which holds all 1024 at a few megabytes each.
Stock keeps 48 resident, which leaves it a cache of \hCapStockFortyEightKv{} tokens, keeps 256 more in host memory, and loads the rest from storage.
Each of its rank-64 conversions is \hBytesSixtyFour\,GB.
A request meets the latency target if its first token arrives within 1\,s and each later token within 100\,ms.
The \emph{sustainable rate} is the highest arrival rate at which at least 90\% of requests meet the target.
Figure~\ref{fig:capacity} shows the share of requests that meet the target at each rate (App.~\ref{app:measure} gives the setup).
Our engine sustains \hCapSixtyFourSustainOurs, \hCapTwoFiftySixSustainOurs{} and \hCapThousandSustainOurs{} requests per second at 64, 256 and 1024 tenants, and stock sustains \hCapThousandSustainStock{} at each, so our sustainable rate is \hCapSustainLo{} to at least \hCapSustainHi{} that of stock.
With 64 tenants, our engine still meets the target at \hCapSixtyFourSustainOurs{} requests per second, the highest rate that we tested, so \hCapSustainHi{} is a lower bound.
At 1024 tenants and 8 requests per second, \hCapThousandSloOurs{} of our requests meet the target, against \hCapThousandSloStock{} for stock.
Our engine delivers \hCapThousandTokOurs{} tokens per second, against \hCapThousandTokStock{} for stock, which is \hCapThousandThr{} the throughput (\hCapThousandThrTen{} at 10 requests per second), and stock's mean TTFT is \hCapThousandTtft{} ours.

\textbf{Because each exact adapter is only a few megabytes, one RTX A6000 serves 1024 tenants at \hCapThousandSustainRatio{} the request rate that their LoRA conversions sustain within the latency target, and at 8 requests per second it delivers twice their throughput.}

\subsection{The A100}
\label{sec:eval:a100}

We repeat the measurements on an A100-SXM4-80GB (App.~\ref{app:a100}), where the mixed batch of all five methods again matches the PEFT library to within \hAhAgreeLo--\hAhAgreeHi{} of each adapter's effect.
With 80\,GB, stock keeps every converted adapter resident, including all twenty at rank 256, and our engine is still faster in every decode-heavy cell.
Stock takes \hAfMixDecLo--\hAfMixDecHi{} our time per output token in the mixed pools at every rate, and \hAfVSixteen{} with 16 VeRA tenants.
The conversions of OFT and (IA)$^3$ take \hAfOSixteen{} and \hAfISixteen{} our time at 16 tenants.
With many tenants, stock keeps 80 conversions resident, and at 64 tenants our engine sustains \hCapASixtyFourSustainOurs{} requests per second against \hCapASixtyFourSustainStock{} for stock.
At 1024 tenants, our engine sustains \hCapAThousandSustainOurs{} requests per second, a rate at which stock meets the target for \hCapAThousandSloLowStock{} of requests.
At 256 tenants, both engines sustain \hCapATwoFiftySixSustainOurs{} requests per second.
At \hCapATopRate{} requests per second, our engine delivers \hCapAThrLo--\hCapAThrHi{} the throughput of stock.
Exact serving costs more on the A100 than on the RTX A6000. In the mixed pools, ours takes \hAfRefLo--\hAfRefHi{} the time of LoRA-only, and under prefill-heavy load stock's mean TTFT is as low as \hAfMixPreLo{} ours (App.~\ref{app:a100}).

\textbf{On the A100, with 80\,GB of memory, our engine still decodes faster than the conversions in every decode-heavy cell.}

\section{Related Work}
\label{sec:related}

\textbf{Serving many adapters.}
PetS~\citep{pets} first served several PEFT methods over one shared encoder model, batching the shared GEMM across tasks and running each task's own operator on its part of the batch.
In our engine, every method's adapter work runs in the same batched kernels as LoRA.
Punica~\citep{punica} and S-LoRA~\citep{slora} introduced the segmented LoRA schedule and adapter paging.
FLoRA~\citep{flora} batches a different LoRA adapter for each example.
Later systems schedule, place, cache or compress LoRA adapters~\citep{lorax,dlora,caraserve,loraserve,chameleon,compressserve}.
DeltaZip~\citep{deltazip} and BitDelta~\citep{bitdelta} serve full fine-tunes by compressing each one's difference from the base model, which, like a conversion to LoRA, is an approximation at a chosen size.
We ask which slots a LoRA schedule needs to serve each method exactly, which slots fold at load, and what each live slot costs.

\textbf{Serving other methods.}
An open vLLM pull request\footnote{\texttt{vllm-project/vllm\#47395}, ``[LoRA] Add DoRA support for serving and linear layers'', open since July 2026.} computes DoRA's scale at load, supports both tensor-parallel layouts, mixes LoRA and DoRA requests, and applies the scale after the LoRA expand.
\citet{scalingdora} fuse DoRA's norm computation into the forward pass of a single adapter, while our engine computes DoRA's scale once at load and so removes that computation from serving.
\citet{oftv2} make OFT training scale by rotating the inputs, and the same identity lets OFT keep the shared GEMM at serving time.
VB-LoRA~\citep{vblora} and Tied-LoRA~\citep{tiedlora} also share factors across layers or adapters and train small parts per adapter.
Where shared factors meet per-adapter vectors, as in VeRA, the shared-factor route applies.

\textbf{Kernels and cost models.}
The expand is a Triton~\citep{triton} kernel in the line of Punica's SGMV and FlashInfer~\citep{flashinfer}.
The byte model is a roofline~\citep{roofline} argument applied to the passes that an adapter adds.

\section{Conclusion}

Serving engines batch requests for many LoRA adapters, but they reject adapters of other PEFT methods, and converting these adapters to LoRA costs accuracy or memory.
This paper shows that modifications to the batched LoRA kernel can be made to facilitate the serving of the main PEFT methods exactly, in one batch, from each adapter's checkpoint alone.

First, we provide a normal form with six slots that expresses LoRA, DoRA, (IA)$^3$, SSF, VeRA and OFT, and a fold rule tells which slots an engine can merge into the adapter at load and which slots must stay live.
The rule shows, for example, that an exact fold of DoRA's scale needs a copy of the base weight for each tenant, so a batched engine must keep that scale live.
Second, we introduce three routes to compute every live slot inside the LoRA schedule, and a byte model that predicts the time that a route adds from the bytes that it moves, with no fitted parameters, where bandwidth sets the cost.
Third, we organize this into a compiler and a run-time patch to vLLM to serve all five methods in one batch, with CUDA graphs and \texttt{torch.compile} on.

As a result, a platform no longer has to choose between the methods that its tenants train and the efficiency of batched serving.
Our engine matches the PEFT library with a mean error of at most \hAgreeWorst{} of each adapter's effect.
Under a burst on an RTX A6000, a mix of four or five methods costs only \hSDecRefLo--\hSDecRefHi{} more decode time than LoRA-only.
Conversion to LoRA, in contrast, loses up to \hConvSvdWorstSixtyFour{} points of accuracy for (IA)$^3$ without the tenant's data, and it gives each VeRA tenant its own copy of the shared factors.
Because each adapter keeps the size of its own parameters, one RTX A6000 serves 64 to 1024 tenants at \hCapSustainLo{} to at least \hCapSustainHi{} the sustainable rate of their conversions.
To serve a new method, an engine only needs to map it to the slots, and the fold rule and the byte model tell in advance which slots stay live and how many bytes they move.
Our measurements use Llama-3.1-8B on RTX A6000 and A100 GPUs, serving costs use synthetic tenants with the shapes of trained adapters, and accuracy uses the fifteen adapters that we trained.

\section*{Acknowledgments}

This work was supported by ONR grant N00014-23-1-2714, DOE grant DE-SC0020345, DOI grant 140D0423C0076, and a Google Cloud Computing Award.

\newpage
\typeout{REFSTART \thepage}
\bibliography{references}
\bibliographystyle{mlsys2025}

\section*{AI Usage Disclosure}
We used Claude Code (Anthropic) to assist with code implementation, benchmarking and plotting scripts, bibliography verification, and drafting and editing the text of the paper under the authors' direction. 
All system designs, mathematical proofs, kernel implementations, experimental results, and text were verified and revised by the authors, who take full responsibility for the content of the paper.

\clearpage
\appendix
\setcounter{proposition}{0}
\setcounter{corollary}{0}
\setcounter{table}{0}
\setcounter{figure}{0}
\setcounter{equation}{0}
\renewcommand{\theproposition}{A\arabic{proposition}}
\renewcommand{\thecorollary}{A\arabic{corollary}}
\renewcommand{\thetable}{A\arabic{table}}
\renewcommand{\thefigure}{A\arabic{figure}}
\renewcommand{\theequation}{A\arabic{equation}}
\renewcommand{\theHproposition}{A\arabic{proposition}}
\renewcommand{\theHcorollary}{A\arabic{corollary}}
\renewcommand{\theHtable}{A\arabic{table}}
\renewcommand{\theHfigure}{A\arabic{figure}}
\renewcommand{\theHequation}{A\arabic{equation}}

\section{Proofs}
\label{app:proofs}

We write the term matrix of adapter $a$ as $\bM = \sum_k c_k\bA_k\,\mathrm{diag}(\bd_k)\,\bB_k\,\mathrm{diag}(\be_k)$.
Then $F_a(\bx) = \bx\bR\bW\,\mathrm{diag}(\bs) + \bb + \bx\bM$ and $\mathrm{rank}\,\bM \le r = \sum_k r_k$.
Two members are equal when they agree for every $\bx$.
This section states the exact versions of the informal results of \S\ref{sec:form} and \S\ref{sec:cost}, each before its proof.

\paragraph{Folds.}
A slot $\bv$ of adapter $a$ that is not the identity \emph{folds} if some member computes $F_a$ for every $\bx$ with $\bv$ set to the identity, and this member differs from $a$ only in tensors that $a$ owns, each with the same shape and ownership.

\begin{proposition}[Fold rule; exact version of Proposition~\ref{prop:fold}]
\label{propA:fold}
Let $r = \sum_k r_k$ be the total rank.
Assume that $4r < \min(\dinn,\dout)$, or, if every $\bA_k$ is shared, that $3r < \dinn$ and $2r < \dout$.
Assume also that no two terms share a factor, that an owned $\bR$ is the adapter's only owned tensor, and that a term with owned $\bd_k$ and $\be_k$ on shared $\bA_k,\bB_k$ is the only term.
Then, for all parameters outside a set of measure zero, a slot folds if and only if the adapter owns a tensor next to it in the product that applies it.
In detail, $c_k$ folds when any of $\bA_k,\bd_k,\bB_k,\be_k$ is owned, $\bd_k$ folds when $\bA_k$ or $\bB_k$ is owned, and $\be_k$ folds when $\bB_k$ is owned, while $\bs$, $\bb$ and $\bR$ never fold.
\end{proposition}

\begin{corollary}[Folding $\bs$; exact version of Corollary~\ref{cor:s}]
\label{corA:s}
If $\bR = \bI$, any member that equals $F_a$ with $\bs = \mathbf{1}$ has a total term rank of at least $\mathrm{rank}(\bW\,\mathrm{diag}(\bs-\mathbf{1})) - r$.
\end{corollary}

Every member of Table~\ref{tab:members} meets the ownership conditions: OFT owns only $\bR$, and VeRA has one term.
The second size condition covers the key and value projections of VeRA, where $\dout = 1024 = 4r$.

\paragraph{Proposition~\ref{propA:fold}, the ``if'' direction.}
Each identity in \S\ref{sec:fold} replaces one owned tensor by a tensor of the same shape and sets the slot to its identity.
For $c_k$, use $c_k\bA_k = (c_k\bA_k)\cdot 1$, and likewise for $\bB_k$, $\bd_k$ and $\be_k$.
For $\bd_k$, use $\bA_k\,\mathrm{diag}(\bd_k)$ or $\mathrm{diag}(\bd_k)\bB_k$.
For $\be_k$, use $\bB_k\,\mathrm{diag}(\be_k)$.
Ownership and shapes do not change, so each is a fold.

\paragraph{Proposition~\ref{propA:fold}, the ``only if'' direction.}
Let $F'$ be a candidate fold of $F_a$ that sets the slot under test to its identity.
It may change every owned tensor, in every term, but no shared tensor.
Equality for all $\bx$ separates the constant and linear parts.
So $\bb' = \bb$, and
\begin{equation}
\bR\bW\mathrm{diag}(\bs) - \bR'\bW\mathrm{diag}(\bs') = \bM' - \bM .
\label{eq:fold-eq}
\end{equation}
The right side has rank at most $2r$.
We call the parameters \emph{generic} when they avoid finitely many proper algebraic subsets of parameter space.
Then every square submatrix of $\bW$ is invertible.
When the $\bA_k$ are shared, the columns of $\bR\bW$ are also in general position with respect to their span.
This excludes a set of measure zero.
The arguments below need nothing else, so they hold for every candidate at once.

\emph{The shift $\bb$.} $\bb$ is the constant part, and nothing else is constant in $\bx$.

\emph{The transform $\bR$.} By assumption, an owned $\bR$ is the only owned tensor.
So $\bs' = \bs$, $\bM' = \bM$, and a fold sets $\bR' = \bI$.
Then \eqref{eq:fold-eq} reads $(\bR-\bI)\bW\,\mathrm{diag}(\bs) = \mathbf{0}$.
$\bR - \bI$ is block-diagonal, so block $j$ gives $(\bR_j - \bI)\bW_j\,\mathrm{diag}(\bs) = \mathbf{0}$, where $\bW_j$ holds the $b_R$ rows of $\bW$ that block $j$ mixes.
For generic $\bW$ and $b_R \le \dout$, $\bW_j$ has full row rank.
So, for $\bs$ without zeros, $\bR_j = \bI$ for every $j$.
The argument also holds when $\dinn > \dout$, where $\bW$ has a left null space.

\emph{The scale $\bs$.} Here $\bR$ is not owned, so $\bR' = \bR$.
The left side of \eqref{eq:fold-eq} is $\bR\bW\,\mathrm{diag}(\bs-\bs')$.
For generic $\bW$ and invertible $\bR$, its rank is $\min(\dinn, |\mathrm{supp}(\bs-\bs')|)$.
So $\bs'$ differs from $\bs$ on at most $2r$ columns.
A fold with $\bs' = \mathbf{1}$ therefore needs $\bs-\mathbf{1}$ to be supported on at most $2r < \dout$ columns, which is false for generic $\bs$.
If every $\bA_k$ is shared, the columns of $\bM$ and $\bM'$ lie in the span of the $\bA_k$, of dimension at most $r$.
The left side spans the $|\mathrm{supp}(\bs-\bs')| \le 2r$ columns of $\bR\bW$ on that support.
For generic $\bW$ and $3r < \dinn$, the two spans meet only in zero, so $\bs' = \bs$.
The same equation gives Corollary~\ref{corA:s}.
A candidate with $\bs' = \mathbf{1}$ has $\mathrm{rank}\,\bM' \ge \mathrm{rank}(\bW\mathrm{diag}(\bs-\mathbf{1})) - \mathrm{rank}\,\bM$.

\emph{The vectors and $c_k$.}
Let $C$ be the span of the coordinate vectors on $\mathrm{supp}(\bs-\bs')$, of dimension at most $2r$.
The left side of \eqref{eq:fold-eq} has its rows in $C$.
So $\bM\bP = \bM'\bP$ for $\bP$ the projection onto the orthogonal complement of $V + C$, for any subspace $V$.
There are finitely many choices of $C$, and $C = \{0\}$ when every $\bA_k$ is shared.
No two terms share a factor.
So, for generic factors and $\dim V + \dim C + r < \dout$, the terms of $\bM$ have independent column spaces and, after $\bP$, independent row spaces.
The rank of a sum of terms after $\bP$ is then the sum of their ranks.
Every $V$ below has dimension at most $r$, so either size condition of the proposition suffices.

\emph{$\be_k$ with $\bB_k$ shared.} Take $V = \mathrm{row}(\bB_k)$.
Term $k$ of $\bM'$ has $\be_k$ set to one, so its rows lie in $V$, and $\bP$ removes it.
Hence $\mathrm{rank}\,\bM'\bP \le r - r_k$.
Term $k$ of $\bM$ has row space $V\mathrm{diag}(\be_k)$.
For $\be_k$ outside a set of measure zero, this space meets $V + C$ and the row spaces of the other terms only in zero.
Hence $\mathrm{rank}\,\bM\bP = r$, a contradiction.
The exceptional set contains more than $\be_k \propto \mathbf{1}$, whose scalar folds into $c_k$.
An $\be_k$ that differs from a constant in only a few entries leaves $V\mathrm{diag}(\be_k) \cap V$ of dimension near $r_k$, and the proposition makes no claim there.

\emph{$c_k$ with nothing owned in term $k$.} Take $V = \{0\}$.
Then $\sum_{j\neq k}\bT_j'\bP = \big(\sum_{j\neq k}\bT_j + (c_k - 1)\bF_k\big)\bP$, where $\bT_j$ is term $j$ and $\bF_k$ is the fixed term $k$ without $c_k$.
The left side has rank at most $r - r_k$.
The right side has rank $r$ whenever $c_k \neq 1$.
So $c_k = 1$.

\emph{$\bd_k$ with $\bA_k$ and $\bB_k$ shared.}
If $\be_k$ is not owned, term $k$ of $\bM - \bM'$ is $\bA_k\,\mathrm{diag}(c_k\bd_k - c_k'\mathbf{1})\,\bB_k\,\mathrm{diag}(\be_k)$.
Its rank is the number of entries of $c_k\bd_k$ that differ from $c_k'$, which is at least one unless $\bd_k\propto\mathbf{1}$.
The argument for $c_k$ then gives a contradiction.
If $\be_k$ is owned and term $k$ is the only term, as in VeRA, then $\bM'\bP = \bM\bP$ with $\bA_k$ of full column rank gives $c_k d_i\,b_{ij}\,e_j = c_k'\,b_{ij}\,e_j'$ for the entries $b_{ij}$ of $\bB_k$ outside $C$.
These entries are all nonzero for generic $\bB_k$, so $d_i$ does not depend on $i$. $\blacksquare$

\paragraph{Stacking.}
For members with $\bR=\bI$, write $F_a(\bx;\bh) = \bh\odot\bs_a + \bb_a + \bx\bM_a$, so that $F_a(\bx) = F_a(\bx;\bx\bW)$.
A stack applies $F_2$ to the output of $F_1$, $(F_2\circ F_1)(\bx) = F_2(\bx; F_1(\bx))$.
Then
\begin{align*}
F_2(\bx;F_1(\bx)) &= (\bx\bW\odot\bs_1 + \bb_1 + \bx\bM_1)\odot\bs_2 \\
&\quad + \bb_2 + \bx\bM_2 \\
&= \bx\bW\odot(\bs_1\odot\bs_2) + (\bb_1\odot\bs_2+\bb_2) \\
&\quad + \bx\bM_1\mathrm{diag}(\bs_2) + \bx\bM_2 .
\end{align*}
$\bM_1\mathrm{diag}(\bs_2)$ is $\bM_1$ with each $\be_j$ replaced by $\be_j\odot\bs_2$, so the stack is a member.
The rewritten $\be_j$ folds into $\bB_j$ when the stack owns a copy of $\bB_j$.
When a server reuses the inner adapter's $\bB_j$ instead of copying it, $\be_j$ is a live slot unless $\bs_2 = \mathbf{1}$.
Exchanging the two members shows that the two orders agree if and only if $\bb_1\odot\bs_2+\bb_2 = \bb_2\odot\bs_1+\bb_1$ and $\bM_1\,\mathrm{diag}(\bs_2-\mathbf{1}) = \bM_2\,\mathrm{diag}(\bs_1-\mathbf{1})$.
If neither member scales, both sides are equal, so stacks of shifts and terms commute.
If one member scales and the other has a shift or a term, the condition fails for generic parameters.

\paragraph{Byte counts.}
The following are the exact versions of Proposition~\ref{prop:cost}.
They count elements read from and written to GPU memory.

\begin{proposition}[Live output slot]
\label{propA:out}
Let a live slot scale a tensor $\bT$ of shape $[N,\dout]$.
If a pass that already reads and writes $\bT$ carries the slot, the slot adds one read of its vector per tile of rows, which for $k$ adapters in the batch is at most $k\dout$ elements.
If the slot runs in a pass of its own, it adds at least $2N\dout$ elements and one launch, and if that pass first writes a copy of the slot for every row, it adds $4N\dout$ elements.
\end{proposition}

\begin{proposition}[Live input slot]
\label{propA:in}
An input transform $\bR$ with blocks of size $b_R$, in a pass of its own, moves $2N\dinn + k\dinn b_R$ elements and performs $b_R/q$ arithmetic operations per byte, where $q$ is the number of bytes per element.
For the PEFT library's default block size of 32, this is far below the point where the RTX A6000 or the A100 becomes limited by arithmetic.
\end{proposition}

\begin{proposition}[Shared factors]
\label{propA:shared}
Let $k$ adapters in a batch each have a term $((\bx\bA)\odot\bd_a)\bB\odot\be_a$ on the same factors $\bA,\bB$ of rank $r$.
Two GEMMs over the batch, with each row's $\bd_a$ and $\be_a$ read alongside, read $r(\dinn+\dout)$ factor elements per step for any $k$, plus $k(r+\dout)$ vector elements.
Converted to LoRA, each adapter owns a copy of the factors, and the batch reads $k\,r(\dinn+\dout)$ elements.
\end{proposition}

\paragraph{Propositions~\ref{propA:out}--\ref{propA:shared}.}
A pass that applies a slot to $\bT$ must read every element of $\bT$ and write every element of the result.
If the pass already reads and writes $\bT$, only the slot's elements are new.
This is one read per row tile, at most $k\dout$ distinct elements from GPU memory, with the rest served from L2.
Otherwise, the read and the write are both new.
Writing a per-row copy of the slot first adds one more $[N,\dout]$ tensor, written and then read.
For $\bR$, the pass reads and writes $[N,\dinn]$ and reads the $\dinn/b_R$ blocks of $b_R^2$ elements of each of $k$ adapters.
Each output element costs $2b_R$ arithmetic operations for $2q$ bytes, which is $b_R/q$ operations per byte with $q$ = itemsize.
For the shared factors, each GEMM over the batch reads its factor once per step, and each row reads its own $\bd_a$ and $\be_a$, which gives $r(\dinn+\dout) + k(r+\dout)$.
A conversion gives each of the $k$ adapters its own pair of rank-$r$ factors, so the batch reads $k\,r(\dinn+\dout)$. $\blacksquare$

\paragraph{Tensor parallelism.}
\label{app:tp}
We show how DoRA's $\bg$ is computed and applied when a layer is split across GPUs.
A column-parallel layer splits $\dout$.
The norm $\lVert\bW+c\bA\bB\rVert_c$ is taken over $\dinn$ for each output column, so each GPU holds complete columns and computes its own part of DoRA's $\bg$.
A row-parallel layer splits $\dinn$.
Each GPU then holds a partial sum of squares for every column, and $\lVert\cdot\rVert_c^2 = \sum_t\lVert\cdot\rVert_{c,t}^2$ needs one all-reduce of $\dout$ values before the square root.
This all-reduce is necessary, because a column norm depends on every part.
At serving time, a row-parallel layer's output is $\sum_t \bh_t$ over GPUs, and $\bs\odot\sum_t(\bh_t + \bV_t\bB') = \sum_t \bs\odot(\bh_t+\bV_t\bB')$.
So each GPU applies $\bs$ in its own expand.
Table~\ref{tab:tpfold} compares each GPU's $\bg$ with the value computed on one GPU, element by element, over \nTpFoldComparisons{} slices.
The largest row-parallel difference is two to four fp32 units in the last place, for values near one.

\begin{table}[ht]
\centering
\caption{DoRA's $\bg$ computed on real tensor-parallel shards against $\bg$ computed on one GPU, four DoRA adapters.}
\label{tab:tpfold}
\small
\begin{tabular}{rlrr}
\toprule
degree & sharding & slices & max $|\Delta\bg|$ \\
\midrule
2 & column-parallel & 512 & 0 \\
2 & row-parallel & 256 & $2.4\times10^{-7}$ \\
4 & column-parallel & 512 & 0 \\
4 & row-parallel & 512 & $2.4\times10^{-7}$ \\
8 & column-parallel & 512 & 0 \\
8 & row-parallel & 1024 & $2.4\times10^{-7}$ \\
\bottomrule
\end{tabular}

\end{table}

\section{Folding DoRA's Scale on Real Weights}
\label{app:foldrank}

We take four public DoRA fine-tunes of Llama-3.1-8B and layers 0, 15 and 31.
For each, we compute $\bg$ exactly as the engine does, and the singular values of $\bW\,\mathrm{diag}(\bg-\mathbf{1})$ in float64 (Table~\ref{tab:foldrank}).
Every one of the \nFoldFullRank{} matrices has full numerical rank.
A trained $\bg$ deviates from one by only \nFoldGdev{} in root mean square.
The deviation is spread over every column, so even the best rank-16 approximation leaves at least \nFoldLeftSixteen{} of the matrix's squared Frobenius norm in every case.

\begin{table}[ht]
\centering
\caption{$\bW\,\mathrm{diag}(\bg-\mathbf{1})$ for each projection, of shape $\dinn\times\dout$: the minimum numerical rank, the minimum share of its squared Frobenius norm outside the best rank-16 approximation, and the median RMS deviation of $\bg$, over four adapters and three layers.}
\label{tab:foldrank}
\small
\begin{tabular}{lrrrr}
\toprule
proj. & shape & rank & left, $r{=}16$ & rms $\bg{-}\mathbf{1}$ \\
\midrule
q & $4096\times4096$ & 4096 & 61.6\% & 0.0022 \\
k & $4096\times1024$ & 1024 & 59.7\% & 0.0019 \\
v & $4096\times1024$ & 1024 & 83.2\% & 0.0031 \\
up & $4096\times14336$ & 4096 & 92.8\% & 0.0023 \\
down & $14336\times4096$ & 4096 & 95.9\% & 0.0018 \\
\bottomrule
\end{tabular}

\end{table}

\section{Converting to LoRA}
\label{app:conversions}

Each conversion approximates the exact update $\Delta\bW$ of every adapted module.
We compute $\Delta\bW$ from the module of the PEFT library, and it matches the module's own output to a relative error of at most $10^{-4}$ in float32.
\emph{SVD} keeps the top $k$ singular directions of $\Delta\bW$.
\emph{ASVD}, activation-aware SVD, minimizes $\lVert\bX(\Delta\bW - \bL)\rVert_F$ over rank-$k$ matrices $\bL$.
Here $\bX$ holds the module's inputs on 128 calibration sequences, captured before the (IA)$^3$ scale or the OFT rotation.
We accumulate the input covariance in float64, regularise it by $10^{-6}$ times its mean diagonal, and take the minimizer in closed form from its eigendecomposition.
The calibration sequences are the first 128 training examples of the adapter's own task (ASVD), of another task (ASVD$_x$: GSM8K from MNLI, MNLI from BoolQ, BoolQ from GSM8K), or 128 chunks of 160 tokens of WikiText-2 (ASVD$_g$).
\emph{Distillation} starts from ASVD.
It minimizes the KL divergence from the next-token distribution of the adapter in the PEFT library on the completion tokens of the adapter's training data, for 300 steps at a learning rate of $10^{-4}$, with the last 200 training examples held out.
We do not distill VeRA, because own-task ASVD at rank 64 is already within 3 points of the PEFT library on every task.
We evaluate conversions with the training harness on the examples of App.~\ref{app:training}.
Its SVD results match the same conversions served by stock vLLM to within 0.4 points.

Table~\ref{tab:conv} gives every conversion at ranks 16 and 64.
Table~\ref{tab:qualityfull} gives every trained adapter, LoRA, DoRA and VeRA included, served by our engine and, converted by SVD, by stock vLLM.

\begin{table*}[ht]
\centering
\caption{Every conversion of the trained (IA)$^3$, OFT and VeRA adapters (VeRA not distilled): SVD, ASVD fitted on the task's own data, on another task's data (x) and on generic text (g), and distillation, evaluated with the training harness. \emph{PEFT} is the accuracy in percent, and the other columns give the difference from PEFT in points.}
\label{tab:conv}
\small
\setlength{\tabcolsep}{3.5pt}
\begin{tabular}{llrrrrrrrrrrr}
\toprule
 & & & \multicolumn{5}{c}{rank 16} & \multicolumn{5}{c}{rank 64} \\
\cmidrule(lr){4-8}\cmidrule(lr){9-13}
method & task & PEFT & SVD & ASVD & ASVD$_x$ & ASVD$_g$ & dist. & SVD & ASVD & ASVD$_x$ & ASVD$_g$ & dist. \\
\midrule
(IA)$^3$ & GSM8K & 59.8 & $-$29.8 & $-$24.6 & $-$34.4 & $-$32.8 & $-$3.2 & $-$18.6 & $-$0.8 & $-$37.8 & $-$19.6 & $-$0.4 \\
 & MNLI & 90.2 & $-$35.9 & $-$0.9 & $-$23.2 & $-$25.6 & +0.3 & $-$23.2 & +0.5 & $-$15.4 & $-$16.9 & $-$0.3 \\
 & BoolQ & 87.9 & $-$10.4 & $-$0.1 & $-$2.1 & $-$4.4 & +0.5 & $-$4.8 & +0.8 & $-$0.3 & $-$1.6 & +0.3 \\
\addlinespace
OFT & GSM8K & 63.0 & $-$23.4 & $-$3.0 & $-$12.6 & $-$16.2 & $-$2.4 & $-$5.6 & $-$1.2 & $-$2.0 & $-$4.6 & $-$2.0 \\
 & MNLI & 90.9 & $-$10.7 & +0.3 & $-$6.5 & $-$9.6 & $-$0.8 & $-$2.9 & +0.3 & $-$0.6 & $-$0.9 & $-$0.1 \\
 & BoolQ & 89.7 & $-$2.3 & $-$0.2 & $-$0.5 & $-$2.5 & $-$0.5 & $-$0.6 & $-$0.1 & 0.0 & $-$0.2 & $-$0.8 \\
\addlinespace
VeRA & GSM8K & 58.4 & $-$14.8 & +2.0 & $-$1.6 & $-$5.4 & --- & +0.8 & $-$0.8 & +3.2 & +3.4 & --- \\
 & MNLI & 90.9 & $-$18.5 & 0.0 & $-$6.8 & $-$10.3 & --- & $-$0.6 & +0.3 & 0.0 & 0.0 & --- \\
 & BoolQ & 89.5 & $-$6.9 & +0.1 & $-$1.0 & $-$1.9 & --- & $-$0.1 & +0.1 & +0.3 & $-$0.1 & --- \\
\bottomrule
\end{tabular}

\end{table*}

\begin{table}[ht]
\centering
\caption{Every trained adapter. \emph{PEFT} is the accuracy in percent when the PEFT library runs the adapter. The other columns give the difference from PEFT in points, in Scute with all fifteen in one batch, and converted to LoRA by SVD and served by stock vLLM. Bold marks the smallest absolute difference in each row. For LoRA, the rank-16 column is the adapter as trained. Stock converts VeRA exactly, at its own rank of 256; Table~\ref{tab:conv} gives its truncated fits. The base model scores \hQBaseGsm, \hQBaseMnli{} and \hQBaseBoolq.}
\label{tab:qualityfull}
\small
\setlength{\tabcolsep}{3.5pt}
\begin{tabular}{llrrrrr}
\toprule
 & & & & \multicolumn{3}{c}{converted to LoRA, rank} \\
\cmidrule(lr){5-7}
method & task & PEFT & Scute & 16 & 64 & 256 \\
\midrule
(IA)$^3$ & GSM8K & 59.8 & \textbf{+0.4} & $-$29.8 & $-$18.2 & $-$6.2 \\
 & MNLI & 90.2 & \textbf{0.0} & $-$35.8 & $-$23.2 & $-$7.1 \\
 & BoolQ & 87.9 & \textbf{{\boldmath$-$}0.1} & $-$10.5 & $-$4.7 & +0.5 \\
\addlinespace
OFT & GSM8K & 63.0 & \textbf{+0.2} & $-$23.4 & $-$5.6 & +0.8 \\
 & MNLI & 90.9 & \textbf{0.0} & $-$10.7 & $-$2.9 & +0.1 \\
 & BoolQ & 89.7 & \textbf{0.0} & $-$2.2 & $-$0.6 & +0.2 \\
\addlinespace
DoRA & GSM8K & 66.0 & \textbf{{\boldmath$-$}0.4} & +1.4 & +1.2 & +1.2 \\
 & MNLI & 91.9 & \textbf{0.0} & \textbf{0.0} & $-$0.1 & \textbf{0.0} \\
 & BoolQ & 88.7 & \textbf{0.0} & \textbf{0.0} & $-$0.1 & \textbf{0.0} \\
\addlinespace
VeRA & GSM8K & 58.4 & \textbf{+0.2} & --- & --- & \textbf{+0.2} \\
 & MNLI & 90.9 & \textbf{+0.1} & --- & --- & \textbf{+0.1} \\
 & BoolQ & 89.5 & \textbf{0.0} & --- & --- & \textbf{0.0} \\
\addlinespace
LoRA & GSM8K & 66.4 & \textbf{+0.2} & \textbf{{\boldmath$-$}0.2} & --- & --- \\
 & MNLI & 90.8 & \textbf{0.0} & \textbf{0.0} & --- & --- \\
 & BoolQ & 90.0 & \textbf{0.0} & \textbf{0.0} & --- & --- \\
\bottomrule
\end{tabular}

\end{table}

\section{Adapters, Training and DoRA in Practice}
\label{app:training}

\paragraph{Training.}
We train fifteen adapters of Llama-3.1-8B with the PEFT library, version 0.20, one per method and task.
LoRA and DoRA use rank 16 and $\alpha = 32$.
VeRA uses rank 256, with its projections saved and the library's default seed.
OFT uses blocks of 32 in the library's default Cayley form.
These four methods adapt $\bq,\bk,\bv,\bo$, gate, up and down.
(IA)$^3$ adapts $\bk$, $\bv$ and the input of the down projection, which is the library's default.
Training uses bf16 autocast, gradient checkpointing, sequences of at most 512 tokens, an effective batch of 16, AdamW without weight decay, and a cosine schedule with 3\% warm-up.
The loss covers the completion only, for $\min(1000,\text{one epoch})$ steps.
For each method, we choose the learning rate with the lowest held-out loss after 100 steps on GSM8K.
GSM8K is scored by exact match of the final number on 500 test problems, and MNLI and BoolQ by accuracy on 1000 validation examples, with greedy decoding and fixed prompts (Table~\ref{tab:training}).

\begin{table}[ht]
\centering
\caption{The trained adapters: learning rate, optimizer steps, and the task metric in the PEFT library.}
\label{tab:training}
\small
\begin{tabular}{llrrr}
\toprule
method & task & lr & steps & metric \\
\midrule
LoRA & GSM8K & 0.0002 & 467 & 0.664 \\
LoRA & MNLI & 0.0002 & 1000 & 0.908 \\
LoRA & BoolQ & 0.0002 & 588 & 0.900 \\
DoRA & GSM8K & 0.0002 & 467 & 0.660 \\
DoRA & MNLI & 0.0002 & 1000 & 0.919 \\
DoRA & BoolQ & 0.0002 & 588 & 0.887 \\
(IA)$^3$ & GSM8K & 0.01 & 467 & 0.598 \\
(IA)$^3$ & MNLI & 0.01 & 1000 & 0.902 \\
(IA)$^3$ & BoolQ & 0.01 & 588 & 0.879 \\
VeRA & GSM8K & 0.01 & 467 & 0.584 \\
VeRA & MNLI & 0.01 & 1000 & 0.909 \\
VeRA & BoolQ & 0.01 & 588 & 0.895 \\
OFT & GSM8K & 0.0001 & 467 & 0.630 \\
OFT & MNLI & 0.0001 & 1000 & 0.909 \\
OFT & BoolQ & 0.0001 & 588 & 0.897 \\
\bottomrule
\end{tabular}

\end{table}

\paragraph{DoRA in practice.}
The trained DoRA adapters have a scale $\bg$ that deviates from one by a median of \hTrainedDoraMedian{} over columns.
We also list every adapter on the Hugging Face Hub that is tagged as an adapter of a Llama-3.1-8B model or that matches a search for DoRA, and read its configuration.
\hSurveyN{} distinct adapters from \hSurveyAuthors{} authors set \texttt{use\_dora}.
For each, we compute $\bg = \bmm\oslash\lVert\bW + c\bA\bB\rVert_c$ from the checkpoint and the base weights in float32.
In \hSurveyBelow{} of them, the median column deviates from one by less than 1\%.
The median adapter's median deviation is \hSurveyMedian{}, and the largest is \hSurveyMax.
This small magnitude is why DoRA converts to LoRA within noise.

\section{Engine Integration}
\label{app:engine}

\paragraph{The patch.}
The integration patches vLLM 0.19.1 at run time, so the installed package stays unchanged.
It patches vLLM's check that rejects DoRA, the tensor-name parser, the tensor loader and the packer, adapter activation and weight creation, and the layer's \texttt{apply}.
The loader and packer carry DoRA's magnitude.
Adapter activation computes DoRA's $\bs$, and weight creation allocates the slot buffers.
The scheduler, paging and eviction, attention and batch composition are unchanged.
The engine computes DoRA's $\bs$ at admission, after it has fused $\bq,\bk,\bv$ and gate, up, and sharded every tensor.
This computation therefore uses the engine's own layout.

\paragraph{CUDA graphs.}
A captured CUDA graph replays the same kernels for any batch.
In an engine that serves OFT or VeRA, every decode step that runs a captured graph therefore runs the split GEMM and the shared term, whatever adapters its batch holds.
Steps outside the captured sizes check the batch on the host, from the adapter mapping that vLLM already builds, and skip these kernels when they are not needed.

\paragraph{Amplified adapters.}
The PEFT library initializes (IA)$^3$, VeRA and OFT at or near the identity.
The agreement pool therefore draws each adapter's vectors so that they move log-probabilities by a clear margin.
For standard normal $\bz$, DoRA uses $\bg = \mathbf{1}+0.3\bz$, (IA)$^3$ uses $\bs = \mathbf{1}+0.3\bz$, VeRA uses $\bd = 0.1(\mathbf{1}+0.3\bz)$ and $\be = 0.5\bz$, and OFT uses skew parameters $0.02\bz$ with blocks of 32.
The pool has two adapters each of DoRA, (IA)$^3$, VeRA and OFT, with different seeds, and one LoRA adapter (Table~\ref{tab:agree}).

\begin{table}[ht]
\centering
\caption{One batch of \hAgreeAdapters{} amplified adapters against the PEFT library in float32. The effect is the mean change in log-probability per token that the adapter, run by the PEFT library, makes to the base model. The error is the engine's largest mean error per method, as a share of that effect, on one GPU and on two GPUs with tensor parallelism.}
\label{tab:agree}
\small
\setlength{\tabcolsep}{4pt}
\begin{tabular}{lrrr}
\toprule
method & effect (nats/token) & error / effect, TP 1 & TP 2 \\
\midrule
LoRA & 0.20 & 1.27\% & 1.51\% \\
DoRA & 1.98 & 0.55\% & 0.47\% \\
(IA)$^3$ & 1.09 & 0.54\% & 0.55\% \\
VeRA & 0.22 & 1.41\% & 1.49\% \\
OFT & 0.86 & 0.34\% & 0.37\% \\
\bottomrule
\end{tabular}

\end{table}

\section{The Byte Model}
\label{app:bytemodel}

\paragraph{Constants.}
$\beta$ is the bandwidth of a 256\,MiB device copy.
$\beta_{L2}$ is the fastest of copies sized to $1/16$, $1/8$ and $1/4$ of the L2 cache.
$\lambda$ is the time per call of 200 one-block kernels chained in one CUDA graph.

\paragraph{Routes.}
We test the model on four routes that apply DoRA's $\bs$ outside the expand, and on OFT's input transform.
A \emph{side kernel} writes $\bh\odot\bs$ before our expand reads it, which adds $2N\dout$ elements and one launch.
A \emph{gathered scale} applies $\bs$ after the expand in one pass, which adds $2N\dout$ elements and one launch.
An \emph{index scale} first writes $\bs$ for every row and then multiplies, which adds $4N\dout$ elements and two launches.
\emph{vLLM outside} is vLLM's expand followed by the open pull request's index, copy and multiply, which adds $4N\dout$ elements and three launches.
It is measured against vLLM's LoRA expand.
Table~\ref{tab:bytemodel} gives the median error of each route.
Below the bandwidth regime, the single constant $\lambda$ stands for each kernel's launch, tail and wait, and the error is larger.

\begin{table}[ht]
\centering
\caption{Median absolute error of the byte model on the RTX A6000. \emph{Increment}: error on the time that the extra passes add. \emph{Total}: error on the route's total time. Both are over the cells where $\Delta Q/\beta > \lambda$. \emph{Launch}: error on total time over the other, launch-bound cells.}
\label{tab:bytemodel}
\small
\setlength{\tabcolsep}{3pt}
\begin{tabular}{lrrrr}
\toprule
route & cells & increment & total & launch \\
\midrule
side kernel & \nBmCellsSide & \nBmIncSide & \nBmErrSide & \nBmErrLaunchSide \\
gathered scale & \nBmCellsGather & \nBmIncGather & \nBmErrGather & \nBmErrLaunchGather \\
index scale & \nBmCellsIndex & \nBmIncIndex & \nBmErrIndex & \nBmErrLaunchIndex \\
vLLM outside & \nBmCellsOutside & \nBmIncOutside & \nBmErrOutside & \nBmErrLaunchOutside \\
OFT transform & \nOftCells & \nOftIncErr & \nOftPredErr & --- \\
\bottomrule
\end{tabular}
\end{table}

\section{Measurement Protocol}
\label{app:measure}

\paragraph{Kernel timing.}
Issuing eight Triton launches from Python costs about 400\,$\mu$s, against roughly 95\,$\mu$s of kernel time.
We therefore time every kernel inside a CUDA graph.
We chain enough copies of a call into one graph that the timed interval reaches 2\,ms, and divide.
We rebuild the graph five times and report the median of the five.

\paragraph{Tuned kernels.}
vLLM ships no LoRA kernel configurations for the RTX A6000.
With its defaults, the fused $\bq\bk\bv$ expand reaches \nDefaultQkvRoofLo--\nDefaultQkvRoofHi{} of copy bandwidth.
We search a grid for vLLM's shrink and expand on every layer shape, choose by LoRA's own time inside a CUDA graph, and write the configuration file that vLLM reads.
Every engine in every serving measurement uses these files, tuned for ranks 16, 32, 64, 128 and 256.
Our slot uses the configurations tuned for LoRA.

\paragraph{Serving.}
Each measurement cell runs one engine in a fresh process.
The engine settings are fixed and recorded: 64 sequences, 8192 batched tokens, a maximum length of 1024, and fp16.
vLLM's on-disk compile cache is off in every engine, because its key does not include run-time patches.
Adapters are admitted before timing.
An untimed warm-up stream of the same shape runs first, because a kernel first compiled inside the timed stream costs about a second of TTFT.
Prompts are random token ids of exact length, so prefix caching never applies.
Arrival times are drawn in advance and submitted on an absolute clock.
Prompts and arrivals change with the repetition, and the engines of one repetition share them.
Over six repetitions, every order of the engines occurs equally often.
Before each cell, its GPUs must be idle in two samples ten seconds apart.
Each cell records its KV-cache capacity and the device memory in use.
Ratios are medians of per-repetition ratios, with percentile-bootstrap intervals over repetitions.
Across \hFOrderCells{} ratios, whether an engine ran first moves the ratio by a median of \hFOrderGap.

\paragraph{TTFT.}
Under a burst, prompts finish prefill in waves of one batched step each.
The median request sits on a wave boundary, so the median TTFT moves by a whole step between repetitions whose fastest, 90th-percentile and slowest requests agree.
We therefore report mean TTFT for prefill-heavy load.

\paragraph{Pools.}
The single-method pools hold 1, 2, 4, 8 and 16 VeRA tenants, 1, 4 and 16 OFT tenants and 1, 4 and 16 (IA)$^3$ tenants, with 128 requests each.
For OFT, a second stock engine serves rank-256 conversions.
The five-method pool holds 20 tenants, four of each method, with 8 requests per tenant.
It runs at decode-heavy rates of 4, 8 and 16 requests per second and burst, and at prefill-heavy rates of 2, 4, 8 and 16 and burst.
The same pool also runs with Zipf popularity and at tensor-parallel degree 2.
The four-method pool holds 16 tenants, four each of LoRA, DoRA, (IA)$^3$ and OFT.
The many-tenant pools hold 64, 256 and 1024 tenants, half (IA)$^3$ and half VeRA.
Each tenant reuses the files of one of 16 synthetic adapters of its method, under its own name and id.
The engine therefore loads, holds and swaps it as a separate adapter.
Each many-tenant cell first admits every tenant with a one-token request.
It then runs an untimed warm-up stream and 512 timed requests, with tenants in turn, and it has three repetitions.
The DoRA comparison of \S\ref{sec:cost} serves four DoRA adapters of rank 16 against four LoRA adapters, at 0.80 of device memory.
Engine threads are capped (\texttt{OMP\_NUM\_THREADS=16}, eight compile threads).

\paragraph{Results.}
Table~\ref{tab:frates} gives every rate of the mixed pools with intervals.
Table~\ref{tab:slopes} sets the decode time that each tenant adds against the byte model.
Table~\ref{tab:decomp} breaks down stock's decode time in the mixed pools.

\begin{table}[ht]
\centering
\caption{Every rate of the mixed pools on the RTX A6000, vLLM as shipped: median per-repetition ratio and 95\% interval. TPOT for decode-heavy load, mean TTFT for prefill-heavy load.}
\label{tab:frates}
\footnotesize
\setlength{\tabcolsep}{2pt}
\begin{tabular}{lllrr}
\toprule
 & & & stock / & Scute / \\
pool & load & req/s & Scute & LoRA-only \\
\midrule
20 tenants & decode & 4 & 1.92 [1.89, 1.94] & 1.13 [1.13, 1.14] \\
 & decode & 8 & 1.82 [1.78, 1.83] & 1.09 [1.06, 1.12] \\
 & decode & 16 & 1.66 [1.65, 1.67] & 1.09 [1.08, 1.09] \\
 & decode & burst & 1.73 [1.73, 1.75] & 1.05 [1.05, 1.06] \\
 & prefill & 2 & 1.36 [1.33, 1.43] & 1.23 [1.15, 1.24] \\
 & prefill & 4 & 1.59 [1.47, 1.71] & 1.29 [1.24, 1.37] \\
 & prefill & 8 & 4.91 [4.29, 5.84] & 1.38 [1.34, 1.46] \\
 & prefill & 16 & 5.96 [4.60, 6.49] & 1.52 [1.46, 1.57] \\
 & prefill & burst & 1.12 [1.11, 1.13] & 1.18 [1.18, 1.19] \\
\addlinespace
16, no VeRA & decode & 4 & 1.21 [1.20, 1.22] & 1.02 [1.02, 1.03] \\
 & decode & 8 & 1.31 [1.30, 1.32] & 0.99 [0.96, 1.02] \\
 & decode & 16 & 1.21 [1.21, 1.22] & 0.97 [0.97, 0.98] \\
 & decode & burst & 1.25 [1.24, 1.25] & 0.94 [0.94, 0.94] \\
 & prefill & 2 & 0.97 [0.95, 0.98] & 1.09 [1.07, 1.11] \\
 & prefill & 4 & 0.97 [0.95, 0.99] & 1.11 [1.09, 1.13] \\
 & prefill & 8 & 1.01 [0.99, 1.06] & 1.14 [1.12, 1.17] \\
 & prefill & 16 & 1.42 [1.27, 1.51] & 1.16 [1.15, 1.16] \\
 & prefill & burst & 0.99 [0.99, 1.00] & 1.05 [1.05, 1.06] \\
\addlinespace
20, Zipf & decode & 8 & 1.98 [1.94, 2.07] & 1.12 [1.07, 1.13] \\
 & decode & burst & 1.82 [1.78, 1.86] & 1.05 [1.05, 1.06] \\
 & prefill & 4 & 1.20 [1.14, 1.22] & 1.18 [1.16, 1.25] \\
 & prefill & burst & 1.07 [1.05, 1.08] & 1.20 [1.19, 1.21] \\
\addlinespace
20, TP 2 & decode & 16 & 1.75 [1.69, 1.76] & 1.26 [1.26, 1.28] \\
 & decode & burst & 1.69 [1.69, 1.69] & 1.24 [1.24, 1.24] \\
 & prefill & burst & 1.23 [1.16, 1.26] & 1.19 [1.17, 1.26] \\
\bottomrule
\end{tabular}

\end{table}

\begin{table}[ht]
\centering
\caption{Decode time added per tenant (ms) on the RTX A6000: the byte model's prediction against the least-squares slope of TPOT over 1--16 tenants. Stock is counted on every LoRA layer at the padded rank, because vLLM runs its kernels there for every adapter.}
\label{tab:slopes}
\small
\setlength{\tabcolsep}{4pt}
\begin{tabular}{llrr}
\toprule
method & arm & predicted & measured \\
\midrule
VeRA & Scute & 0.00 & 0.05 \\
 & rank 256 & 1.97 & 1.94 \\
\addlinespace
OFT & Scute & 0.12 & 0.16 \\
 & rank 64 & 0.49 & 0.53 \\
 & rank 256 & 1.97 & 1.94 \\
\addlinespace
(IA)$^3$ & Scute & 0.00 & 0.12 \\
 & rank 64 & 0.49 & 0.53 \\
\bottomrule
\end{tabular}

\end{table}

\begin{table}[ht]
\centering
\caption{Decode time over the base model under a decode burst (ms), vLLM as shipped. The byte model's account of stock, built up step by step, against stock's measured time and Scute's.}
\label{tab:decomp}
\small
\setlength{\tabcolsep}{3pt}
\begin{tabular}{lrr}
\toprule
 & 20 tenants & 16, no VeRA \\
\midrule
conversions at own rank & 13.0 & 5.1 \\
\quad + zero factors elsewhere & 14.3 & 6.4 \\
\quad + padding (vLLM) & 31.6 & 7.9 \\
stock, measured & 34.7 & 10.5 \\
Scute, predicted & 3.1 & 1.5 \\
Scute, measured & 7.1 & 2.4 \\
\bottomrule
\end{tabular}

\end{table}

\section{The A100}
\label{app:a100}

Each run uses one A100-SXM4-80GB and the same source code as the RTX A6000 runs.
vLLM's kernels are tuned on the A100 for ranks 16, 64 and 256.
Each cell has six repetitions, with every engine order equally often.
Stock runs at 0.90 of memory with every adapter resident.
Ours runs at 0.86, which on 80\,GB uses about the same device memory.
Table~\ref{tab:a100} gives the single-method pools and the mixed pools under a burst.
Table~\ref{tab:a100mix} gives every rate of the mixed pools.

For many tenants, stock keeps 80 rank-64 conversions resident and 256 in host memory, and ours keeps 64 tenants resident.
The rates are scaled to the A100's shorter step (Table~\ref{tab:capa100}).

\begin{table}[ht]
\centering
\caption{The A100: median per-repetition ratio under a decode burst (TPOT) and a prefill burst (mean TTFT), stock with every adapter resident.}
\label{tab:a100}
\small
\setlength{\tabcolsep}{3.5pt}
\begin{tabular}{lllrr}
\toprule
 & & & \multicolumn{2}{c}{burst} \\
\cmidrule(lr){4-5}
method & tenants & ratio & TPOT & TTFT \\
\midrule
VeRA & 1 & stock / Scute & 1.06 & 1.04 \\
 & 4 & stock / Scute & 1.15 & 1.05 \\
 & 16 & stock / Scute & 1.83 & 1.10 \\
\addlinespace
OFT & 1 & rank 64 / Scute & 1.07 & 1.00 \\
 & 1 & rank 256 / Scute & 1.23 & 1.13 \\
 & 16 & rank 64 / Scute & 1.31 & 1.04 \\
 & 16 & rank 256 / Scute & 1.99 & 1.21 \\
\addlinespace
(IA)$^3$ & 1 & stock / Scute & 1.11 & 1.04 \\
 & 16 & stock / Scute & 1.34 & 1.05 \\
\addlinespace
mix & 20 & stock / Scute & 1.63 & 1.00 \\
 & & Scute / LoRA-only & 1.23 & 1.24 \\
mix, no VeRA & 16 & stock / Scute & 1.14 & 0.96 \\
 & & Scute / LoRA-only & 1.07 & 1.08 \\
\bottomrule
\end{tabular}

\end{table}

\begin{table}[ht]
\centering
\caption{The mixed pools on the A100: median per-repetition ratio and 95\% interval. TPOT for decode-heavy load, mean TTFT for prefill-heavy load.}
\label{tab:a100mix}
\footnotesize
\setlength{\tabcolsep}{2pt}
\begin{tabular}{lllrr}
\toprule
 & & & stock / & Scute / \\
pool & load & req/s & Scute & LoRA-only \\
\midrule
20 tenants & decode & 8 & 1.69 [1.68, 1.71] & 1.34 [1.28, 1.36] \\
 & decode & 16 & 1.69 [1.64, 1.70] & 1.26 [1.25, 1.30] \\
 & decode & burst & 1.63 [1.62, 1.63] & 1.23 [1.23, 1.23] \\
 & prefill & 4 & 0.77 [0.70, 0.81] & 1.38 [1.32, 1.65] \\
 & prefill & 16 & 0.95 [0.84, 1.43] & 1.89 [1.81, 2.02] \\
 & prefill & burst & 1.00 [1.00, 1.01] & 1.24 [1.23, 1.27] \\
\addlinespace
16, no VeRA & decode & 8 & 1.12 [1.11, 1.12] & 1.16 [1.14, 1.18] \\
 & decode & 16 & 1.13 [1.12, 1.14] & 1.10 [1.09, 1.10] \\
 & decode & burst & 1.14 [1.14, 1.14] & 1.07 [1.07, 1.08] \\
 & prefill & 4 & 0.81 [0.79, 0.87] & 1.26 [1.19, 1.27] \\
 & prefill & 16 & 0.76 [0.73, 0.83] & 1.45 [1.38, 1.47] \\
 & prefill & burst & 0.96 [0.96, 0.97] & 1.08 [1.07, 1.09] \\
\bottomrule
\end{tabular}

\end{table}

\begin{table}[ht]
\centering
\caption{The A100: requests that meet the latency target (\%) against decode-heavy arrival rate (requests per second), median of three repetitions.}
\label{tab:capa100}
\small
\setlength{\tabcolsep}{3.5pt}
\begin{tabular}{lrrrrr}
\toprule
tenants & 8 & 12 & 16 & 20 & 24 \\
\midrule
64 Scute & 100 & 100 & 100 & 100 & 47 \\
 stock & 100 & 83 & 21 & 14 & 12 \\
\addlinespace
256 Scute & 100 & 70 & 12 & 12 & 12 \\
 stock & 100 & 24 & 13 & 12 & 12 \\
\addlinespace
1024 Scute & 100 & 33 & 12 & 12 & 9 \\
 stock & 32 & 10 & 4 & 2 & 1 \\
\bottomrule
\end{tabular}

\end{table}

\end{document}